\documentclass[11pt]{article}

\usepackage[preprint]{acl}

\usepackage{times}
\usepackage{latexsym}

\usepackage[T1]{fontenc}

\usepackage[utf8]{inputenc}

\usepackage{microtype}

\usepackage{inconsolata}

\usepackage{graphicx}
\usepackage{bbm}
\usepackage{amsmath} 
\usepackage{booktabs} 
\usepackage{subcaption}
\usepackage{multirow}
\usepackage{cleveref}
\usepackage{tikz}
\usepackage{forest}

\tikzset{
    mynode/.style={
        rectangle,
        rounded corners=3pt,
        draw=black, 
        thick,
        inner sep=6pt,
        align=center
    }
}

\title{Leveraging Speech Acts for Low-Data and Cross-Domain Conversation Derailment Forecasting}

\author{Angela Yifei Yuan \\
  The University of Melbourne \\
  \texttt{yuanay@student.unimelb.edu.au} \\
  \And
  Christine De Kock \\
  The University of Melbourne \\
  \texttt{christine.dekock@unimelb.edu.au} \\
  \AND
  Christopher Leckie \\
  The University of Melbourne \\
  \texttt{caleckie@unimelb.edu.au} \\}

\begin{document}
\maketitle
\begin{abstract}
Conversational derailment forecasting aims to predict when online discussions will escalate into hostility, enabling proactive moderation. 
Existing approaches often struggle in low-data settings and to generalize across domains. 
This poses a challenge for new platforms and smaller communities where annotated data is limited. 
We propose modeling pragmatic representations of conversations to reduce lexical noise and improve generalizability. Specifically, speech act information is used as an auxiliary learning signal alongside textual semantics. Experimental results show improved performance across three datasets, particularly in low-data and cross-domain settings.
\end{abstract}

\section{Introduction}
Humans can often anticipate when a disagreement is likely to escalate into hostility, allowing them to intervene before conversations deteriorate. Conversational derailment forecasting aims to computationally replicate this ability by predicting hostile breakdowns before they occur, assisting with proactive online platform moderation~\cite{Evaluate-CMVLarge-Tran2025}. 

Due to the subtle and context-dependent nature of conversational derailment across diverse online communities, computational modeling of this process presents a significant challenge. The language used in a political debate on Reddit looks vastly different from a technical GitHub conversation, and the specific vocabulary of a heated dispute can further diverge. Existing forecasting systems~\cite{CRAFT-CMV-Chang2019, dynamic-Kementchedjhieva2021} primarily rely on semantic features and frequently struggle to adapt to these lexical shifts, or require large volumes of data to learn broadly generalizable features. However, data annotation is expensive, and labelled data is often limited for new platforms and smaller communities. This creates a need for models that perform reliably in low-data settings, including by leveraging transferable signals from data-rich platforms.

To address these challenges, we propose a novel approach to model speech acts (SA) to inform derailment forecasting. Grounded in Speech Act Theory~\cite{SpeechActTheory-Austin1975}, we hypothesize that SAs provide a low-dimensional pragmatic representation that mitigates lexical noise and improves forecasting generalizability. 
To effectively fuse pragmatic and semantic signals without inference latency, we introduce two architectures that integrate SAs via an auxiliary task. 
Unlike recent approaches that rely on domain-specific features and inference-time auxiliary information~\cite{GraphCN-Altarawneh2023,Summary-Hua2024,LLMDerail-Zhang2025}, we use transferable pragmatic tracking to address the gap in low-data and cross-domain dynamic derailment forecasting.

Our contributions are as follows:
\begin{enumerate}
    \item We propose a novel approach for incorporating SA information into derailment forecasting. Extensive experiments across three datasets demonstrate improved generalizability over state-of-the-art baselines, particularly in low-data and cross-dataset settings;
    \item We investigate the relationship between dataset divergence and the cross-dataset generalizability of models based on textual and pragmatic features, showing that pragmatic features yield greater benefits when transferring to highly divergent datasets; and
    \item We introduce an evaluation protocol for dynamic forecasting performance through average temporal aggregation, better reflecting models' per-timestep behavior.
\end{enumerate}

\section{Related Work}
\subsection{Conversational Derailment Forecasting}

Conversational derailment forecasting aims to predict if an initially civil conversation will eventually degrade into hostile behavior~\cite{Wiki-Zhang2018}. Unlike standard toxicity detection, which classifies the abusiveness of utterances after they have occurred~\cite{hateSpeechDetection-Poletto2020}, derailment forecasting dynamically anticipates them, enabling moderator intervention before discussions get out of control.

Prior works on conversational derailment forecasting have proposed varied approaches, ranging from traditional machine learning models with handcrafted linguistic features~\cite{Wiki-Zhang2018} to deep learning architectures operating on raw text and conversational metadata~\cite{CRAFT-CMV-Chang2019, GraphCN-Altarawneh2023}. More recently, pretrained language models (PLMs) have been widely adopted, driven by their pretrained generalizable language representations that transfer well to conversational forecasting tasks~\cite{dynamic-Kementchedjhieva2021,Summary-Hua2024,Evaluate-CMVLarge-Tran2025}. 
The in-context learning capabilities of large language models (LLMs) has also been assessed via few-shot prompting~\cite{Github-Imran2025,LLMDerail-Zhang2025}. 
While LLMs introduce high computational costs and inference latency~\cite{LLMDerail-Zhang2025}, making them unsuitable for real-time moderation on high-traffic platforms, prior work has primarily evaluated them using conversation summaries or full dialogues, rather than dynamic, turn-by-turn forecasting.

Current approaches leverage available datasets to achieve strong supervised performance, and largely ignore the challenges of low-data regimes and cross-domain generalizability. Collecting derailment datasets is resource-intensive, requiring extensive manual annotation~\cite{Wiki-Zhang2018} in addition to explicit moderator actions (e.g., thread locking, comment removal) for initial filtering~\cite{CRAFT-CMV-Chang2019, Github-Imran2025}.
In practice, new platforms, niche communities, or domain-specific conversational settings often possess limited annotated data. Adapting derailment forecasting models to these resource-constrained or out-of-domain environments remains heavily underexplored, highlighting a need for models capable of cross-domain transfer without relying on large platform-specific datasets.

\subsection{Speech Act Modeling}
Speech Act Theory posits that language performs actions beyond conveying information~\cite{SpeechActTheory-Austin1975}. SA verbs (e.g., \textit{complain}) categorize these actions, encapsulating nuanced speaker intents and mental states~\cite{SAMeaning-Goddard2013}.

In natural language processing, mapping high-dimensional text to a low-dimensional action space is often used for two purposes: reducing domain-specific lexical noise and capturing the latent pragmatic functions of utterances. In task-oriented dialogue systems, translating user text into a set of dialogue acts is frequently adopted as the first step for capturing user intent, prior to system response generation~\cite{TOD-Wu2020}. Explicit SA-based clustering has also been shown to improve cross-domain generalizability in explainable politeness detection by reducing lexical reliance~\cite{SAGeneralizability-Aljanaideh2025}. Process mining of digital corpora further identifies patterns in SA sequences~\cite{SAOpenDomain-Compagno2018}, highlighting their potential to serve as sequential signals for conversational forecasting.

Effectively leveraging these pragmatic properties for real-time forecasting presents architectural challenges, which this work seeks to address: balancing SA and textual features, and avoiding the inference latency from real-time SA extraction.

\section{Proposed Approach}
In this section, we describe our proposed approach to (1) extract SA information from raw dialogues, and (2) design derailment forecasting models that utilize the extracted SA information.

\subsection{Speech Act Extraction}\label{sec:Approach-SAExtraction}
\paragraph {Taxonomy construction} We employ a curated SA taxonomy consisting of 50 SA verbs frequently utilized in linguistic and conversational analysis literature~\cite{SAMeaning-Goddard2013, SAOpenDomain-Compagno2018, SAUsage-Isnaeni2025, SAUsage-Kholid2024, SATaxonomy-Searle1975} and an additional ``other'' label. Each SA is accompanied by a brief description of its meaning and, where applicable, the speaker’s psychological state, derived from semantic linguistic works and standard lexicons~\cite{SAMeaning-Vanderveken1990, SAMeaning-Goddard2013, SAMeaning-Wierzbicka1987, CambridgeDictionary}. The taxonomy is detailed in Appendix~\ref{sec:appendix-SA-extraction}.

\paragraph{Zero-shot inference} Using this taxonomy, we prompt the open-source \texttt{gpt-oss-120b}~\cite{gptoss120b} to derive auxiliary SA information. 
LLM-based auxiliary knowledge, such as forecasted next utterances or emotional trajectories, has shown effectiveness in conversational forecasting tasks~\cite{DialogGLP-Wang2023, LLMDerail-Zhang2025}. 
Following existing methodologies, we formulate this as a sentence-level multi-label classification task~\cite{SAOpenDomain-Compagno2018}. An utterance is assigned any speech act that appears in at least one of its constituent sentences.

\paragraph{Prompt development} Prior work on pragmatic analysis and SA extraction explore discrepancies between literal phrasing and underlying intention~\cite{Diplomat-Li2023, RSA-spinosoDiPiano2025, Indirectness-Orsini2025}, while research in derailment forecasting shows that conversational outcomes are further influenced by listener's perception of speakers' intents~\cite{IntentPerception-Chang2020}. 
To address these dimensions, we designed our prompt to target intended SAs as perceived by listeners. Drawing on pragmatic analysis research, the prompt (shown in Table~\ref{tab:sa-extract-prompt}) explicitly instructs the model to identify factors that cause literal and intended meanings to diverge, including figurative language~\cite{RSA-spinosoDiPiano2025, Diplomat-Li2023}, covert aggression~\cite{CyberAggression-Ollagnier2024}, indirectness~\cite{Indirectness-Orsini2025} and violation of conversational maxims~\cite{Diplomat-Li2023}. 
For example, the utterance ``Yeah, you're right, you should definitely be able to redefine terms like activity and external harms.'' may be literally interpreted as ``agreeing'' with the listener. However, within a contentious exchange, listeners would likely perceive the speaker’s intent as the extracted SAs of ``disagreeing'' and ``criticizing''.

\paragraph{Human validation} While the primary objective of our framework is downstream forecasting performance, we evaluated whether LLM-extracted SAs align with human judgments to provide a meaningful pragmatic signal. Specifically, 85 sentences from 30 randomly selected responses were manually labeled. Given the inherent subjectivity of speech act annotation, the LLM demonstrated a moderate Cohen's Kappa agreement ($\kappa = 0.53$) across the 51 granular classes, and almost perfect agreement ($\kappa = 0.88$) when collapsed into 5 fundamental categories (assertive, directive, commissive, expressive, and `other' replacing declarative). This is consistent with the human inter-annotator agreement observed for SA annotation by \citet{SAOpenDomain-Compagno2018}. Further details on the SA extraction evaluation are available in Appendix~\ref{sec:appendix-SA-extraction}.

\subsection{Derailment Forecasting Model}\label{sec:Approach-Model}
The extracted SA information is used via an auxiliary SA detection task during model training to regularize the model's latent space with pragmatic signals. Because this information is strictly a training signal, SA extraction is not required at inference time, eliminating both the cost and latency of running external extraction pipelines.

\begin{figure}[t]
    \centering

    \begin{subfigure}[b]{0.544\linewidth}
        \centering
        \includegraphics[width=1.02\linewidth]{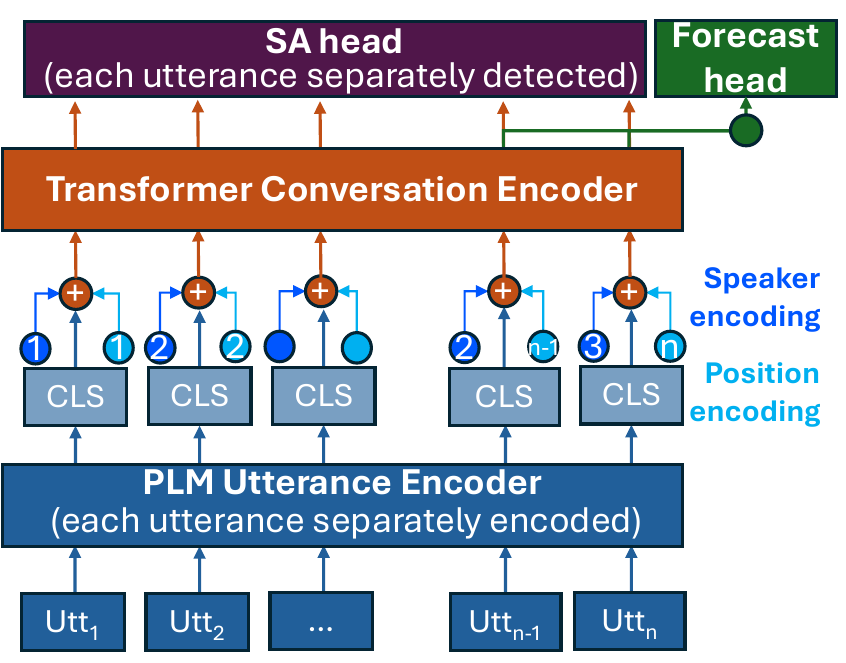}
        \caption{$H_{\text{parallel}}$}
        \label{fig:H-parallel}
    \end{subfigure}
    \hfill
    \begin{subfigure}[b]{0.444\linewidth}
        \centering
        \includegraphics[width=1.02\linewidth]{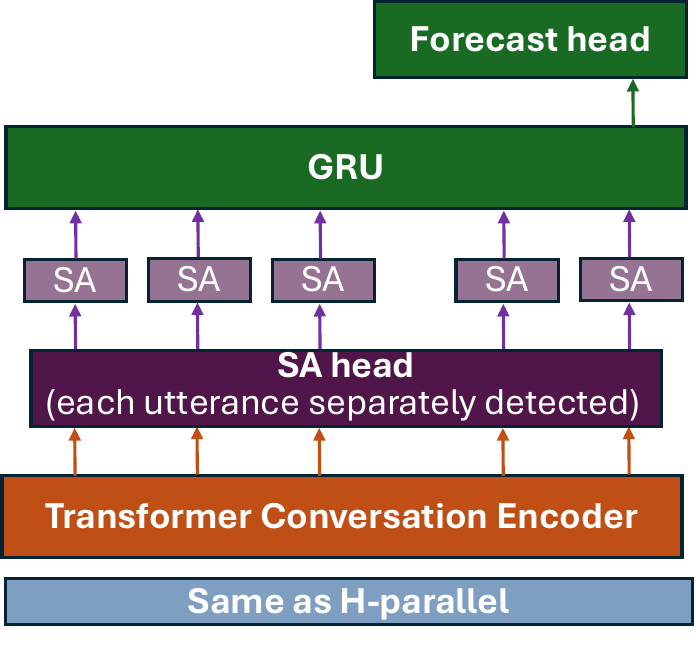}
        \caption{$H_{\text{sequential}}$}
        \label{fig:H-sequential}
    \end{subfigure}
    \caption{Architecture of $H_{\text{parallel}}$ and $H_{\text{sequential}}$.}
    \label{fig:model}
\end{figure}

We designed two models that incorporate both raw text and SA information, with one specifically configured to prioritize SA signals. As illustrated in Figure~\ref{fig:model}, both utilize a hierarchical architecture, which has proven effective in modeling dynamics for conversational prediction tasks~\cite{CRAFT-CMV-Chang2019, TransformerHierarchical-Yuan2023}, and also naturally facilitates the integration of SA labels at the utterance level.
The architecture is composed of two primary components. Firstly, a PLM-based utterance-level encoder (e.g., \texttt{RoBERTa-Large};~\citealp{roberta-Liu2018}) processes the raw text of individual utterances and generates an embedding for each. Secondly, a two-layer Transformer-based conversation-level encoder takes PLM-encoded utterance embeddings, together with speaker and utterance ID embeddings, and outputs contextualized representations for each utterance, capturing its conversational context.

On top of these components, two classification heads are appended for SA detection ($Head_{SA}$) and derailment forecasting ($Head_{derail}$) respectively. $Head_{SA}$ is applied to each contextualized utterance embedding for per-utterance multi-label SA classification. It leverages the hierarchical architecture to prioritize the target utterance while attending to the contextual information.

The two proposed models differ in the design of $Head_{derail}$. 
In $H_{\text{parallel}}$ (Figure~\ref{fig:H-parallel}), $Head_{derail}$ is attached on top of the conversation-level encoder in parallel with $Head_{SA}$. It averages the embeddings from the final turn of conversation to produce the prediction. 
To avoid the model relying mainly on raw textual features instead of SA information, $H_{\text{parallel}}$ undergoes two-phased training: an initial intermediate phase focused solely on SA detection, followed by multi-task training for derailment forecasting with SA detection as an auxiliary task.
The second model, $H_{\text{sequential}}$ (Figure~\ref{fig:H-sequential}), is designed to prioritize SA signal over text. A uni-directional GRU layer~\cite{GRU-Cho2014} is built upon the SA probabilities from $Head_{SA}$. $Head_{derail}$ then predicts based on the final GRU state. By relying primarily on the detected SA probabilities for forecasting, the model directly performs multi-task training without an intermediate SA-only phase.

\section{Experimental setup}
\subsection{Derailment Datasets}\label{sec:dataset}
We consider three datasets covering conversational derailment across different platforms, representing a broad spectrum of online conversations: editorial negotiations (Wikipedia); open-domain, inherently argumentative debates (Reddit ChangeMyView); and goal-oriented technical discussions (GitHub). This diversity enables us to evaluate the models' cross-domain generalizability.

\paragraph{Wikipedia data (WIKI)}~\cite{Wiki-Zhang2018, CRAFT-CMV-Chang2019}: Conversations were collected from publicly accessible talk pages for Wikipedia editors and labeled by crowdworkers as either containing a personal attack toward another user or remaining civil throughout. Each derailed conversation is paired with a civil conversation from the same talk page to avoid topic-specific trivial correlations. WIKI contains 4,188 conversations, partitioned into a train-validation-test split of 60-20-20. Paired conversations are kept within the same split to preserve topic control.

\paragraph{Reddit CMV data (CMV)}~\cite{CRAFT-CMV-Chang2019, Evaluate-CMVLarge-Tran2025}: Dialogues were collected from the subreddit ChangeMyView and labeled based on whether a conversation eventually had a comment removed by a moderator for violating Rule 2: ``Don’t be rude or hostile to other users.'' Similar to WIKI, each derailed dialogue is paired with a civil dialogue that belongs to the same top-level post. CMV contains 19,578 conversations, split similarly to WIKI.

\paragraph{GitHub data (GITHUB)}~\cite{Github-Imran2025}: Conversations were collected from GitHub issues and pull requests. Toxic conversations were primarily sourced from threads locked by moderators as ``too heated'', ``spam'', or ``off-topic'', with individual toxic comments identified via a combination of \texttt{GPT-4o} and human verification. Civil conversations were sampled from the same repositories with a ratio of four civil threads per toxic thread, subject to repository availability, to better reflect a realistic data distribution. This yielded 202 derailed and 696 civil dialogues. Following WIKI and CMV, we split the dataset using a 60-20-20 ratio.

All datasets are preprocessed to only contain responses prior to derailment, and civil dialogues in the training sets are truncated accordingly to prevent trivial correlations related to dialogue length. Since the GITHUB dataset captures entire thread discussion and can be exceptionally long (e.g., 120 utterances), we crop all GITHUB dialogues to their 10 most recent utterances. After processing, all training datasets have a median length of 5 utterances, with averages of 6.02 (CMV), 5.28 (WIKI), and 5.24 (GITHUB) utterances, respectively.

SAs of 51 classes were extracted for training sets following Section~\ref{sec:Approach-SAExtraction}. To evaluate robustness of our models to the SAs used, Appendix~\ref{sec:appendix-SA-robustness} details supplementary experiments on a coarser 18-class SA taxonomy, demonstrating similar performance.

\subsection{Evaluation Task and Metrics}
\paragraph{Dynamic Evaluation} In alignment with prior work~\cite{CRAFT-CMV-Chang2019, Evaluate-CMVLarge-Tran2025}, models are trained statically on full preprocessed dialogues with derailed utterances removed, and evaluated on their dynamic forecasting performance. Let $C=(u^1,u^2, \cdots,u^{n})$ be any conversation consisting of $n$ utterances. During training, models receive entire $C$ and their corresponding labels. During evaluation, models make a prediction $p^t$ given each of $C^t=(u^1,u^2, \cdots,u^{t})$, for $t\in[1,n]$, generating a sequence of $n$ probabilities $p=(p^1,p^2,\cdots,p^{n})$. 

\paragraph{Temporal Aggregation} To evaluate this dynamic performance using standard classification metrics such as the F1 score, the utterance-level predictions are aggregated into a single conversation-level forecast and compared against the ground-truth label. Prior work employs max aggregation ($\text{Dynamic}_{\text{max}}$), converting the highest probability into a binary label via a threshold tuned on the validation set (i.e., $\hat{y}=\mathbbm{1}_{\{max(p) > t_{max}\}}$)~\cite{CRAFT-CMV-Chang2019}. This implies that a conversation is predicted to derail if any dynamic prediction crosses the threshold, making the evaluation highly sensitive to false positive spikes (e.g., a civil dialogue being flagged as derailed due to a single anomalous prediction at turn two, despite recovering to low probabilities for the remainder of the sequence). To address this, we perform mean aggregation ($\text{Dynamic}_{\text{mean}}$, i.e., $\hat{y}=\mathbbm{1}_{\{mean(p) > t_{mean}\}}$), which remains high only if the model signals derailment early and consistently in the exchange, rather than spiking solely at specific turns or toward the final utterance. Furthermore, to avoid the loss of temporal granularity caused by aggregation, we explicitly examine model performance at individual timesteps as dialogues approach the target events (derailments or end of civil dialogues). Specifically, we evaluate the models' predictions exactly $k$ steps away from the event, where $k\in[1,5]$.

\paragraph{Metrics} For evaluation metrics, we report the Area Under the Precision-Recall Curve (AUPRC) and Macro-F1 score, both of which are robust to imbalanced datasets (such as the GITHUB corpus). AUPRC avoids the need for threshold tuning by assessing the model's ability to rank positive cases above negative cases across all thresholds. This metric is effective for evaluating generalizability to OOD settings (where models trained on one dataset are tested on the others), as it isolates the model's capacity to provide derailment signals from its reliance on domain-specific threshold calibration. Conversely, the Macro-F1 score provides a complementary evaluation of the model's practical performance when deployed at a specific threshold.

\paragraph{Training Set Sizes} To assess model performance across varying data regimes, we train the models using different amounts of training data and report the Area Under the Learning Curve (AULC;~\citealp{AULC-Viering2023}). Specifically, we utilize training subset sizes of 300, 500, 1000, 2500, and 11802, capped by the maximum capacity of each respective dataset. These subset sizes allow us to evaluate how model performance scales with increasing data up to approximately the full availability of the training corpora. The full training sets contain 2508 (WIKI), 538 (GITHUB), and 11802 (CMV) dialogues. The AULC is calculated as the integral of the performance metrics (AUPRC and Macro-F1) plotted against the log-scaled training size. To facilitate easier interpretation across datasets, the AULC is normalized to a range of 0 to 1; this linear scaling preserves the relative performance differences and does not affect model comparisons.

\begin{table*}[t]
\centering
\small
\setlength{\tabcolsep}{6pt}
\begin{tabular}{lcccccc}
\toprule
& \multicolumn{2}{c}{WIKI} & \multicolumn{2}{c}{CMV} & \multicolumn{2}{c}{GITHUB} \\
\cmidrule(lr){2-3} \cmidrule(lr){4-5} \cmidrule(lr){6-7}
Model 
& MacroF1 & AUPRC
& MacroF1 & AUPRC
& MacroF1 & AUPRC \\
\midrule
CRAFT\textsuperscript{\ref{fn:CRAFT}}
& $60.97 \pm 0.74^\dagger$ & $63.59 \pm 0.69^\dagger$
& $61.97 \pm 0.47^\dagger$ & $62.76 \pm 0.41^\dagger$
& -- & -- \\

PLM
& \underline{$61.77 \pm 0.69$} & $64.09 \pm 0.96^\dagger$
& \underline{$65.11 \pm 0.61$} & $66.72 \pm 0.42^\dagger$
& \underline{$71.06 \pm 1.92$} & $57.16 \pm 4.56^\dagger$ \\

$H_{\text{ablation}}$
& $61.03 \pm 0.68^\dagger$ & \underline{$65.35 \pm 1.10$}
& $64.89 \pm 0.38$ & \underline{$66.90 \pm 0.61^\dagger$}
& $66.50 \pm 1.59^\dagger$ & \underline{$58.49 \pm 4.52^\dagger$} \\
\hline

$H_{\text{sequential}}$
& \underline{$61.77 \pm 0.75$} & $62.82 \pm 0.83^\dagger$
& $63.07 \pm 0.51^\dagger$ & $64.95 \pm 0.57^\dagger$
& $70.46 \pm 1.60$ & $54.76 \pm 2.83^\dagger$ \\

$H_{\text{parallel}}$
& $\mathbf{62.50 \pm 0.98}$ & $\mathbf{66.85 \pm 0.71}$
& $\mathbf{65.29 \pm 0.45}$ & $\mathbf{67.94 \pm 0.26}$
& $\mathbf{71.55 \pm 3.47}$ & $\mathbf{66.84 \pm 1.65}$ \\

\bottomrule
\end{tabular}
\caption{Comparison of models' $\text{Dynamic}_{\text{mean}}$ AULC performance. Best results are \textbf{bolded} and second-best are \underline{underlined}. The $\dagger$ symbol denotes that the model's performance is statistically significantly different to the best model ($p < 0.05$, two-tailed paired t-test). $H_{\text{parallel}}$ achieves the strongest performance across the datasets.}
\label{tab:in-domain-mean-AULC}
\end{table*}

\subsection{Baseline Models}
We compare $H_{\text{parallel}}$ and $H_{\text{sequential}}$ against SOTA derailment forecasting models. CRAFT~\cite{CRAFT-CMV-Chang2019} was introduced as the first dynamic derailment forecasting model. It employs a hierarchical recurrent encoder-decoder (HRED) architecture that learns to represent conversational dynamics through unsupervised pretraining on a large corpus via a next-response generation objective, before being fine-tuned for the conversational derailment task. To benefit from larger-scale general pretraining, recent approaches~\cite{dynamic-Kementchedjhieva2021, TransformerHierarchical-Yuan2023, Evaluate-CMVLarge-Tran2025} directly employ PLMs. Among these, \texttt{RoBERTa-Large}~\cite{roberta-Liu2018} and \texttt{BERT-Base}~\cite{BERT-Devlin2018} have been widely adopted for conversational forecasting tasks, including predicting derailment~\cite{Evaluate-CMVLarge-Tran2025,dynamic-Kementchedjhieva2021} and emotion~\cite{EmotionACEIC-Xu2025, EmotionDialogueGLP-Wang2023}. Given its larger architecture and superior performance, \texttt{RoBERTa-Large} is selected as both a SOTA baseline (denoted as PLM in tables and figures) and the utterance encoder for our proposed models. 
Additionally, as both a baseline and an ablation study, we evaluate $H_{\text{ablation}}$, which shares the same hierarchical architecture of $H_{\text{parallel}}$ adapted from~\cite{TransformerHierarchical-Yuan2023}, but is trained without incorporating SA knowledge. 
To evaluate the robustness of the proposed method to the choice of underlying PLM and model architecture, two supplementary experiments utilizing BERT and a graph-based model GraphNLI~\cite{GraphNLI-Agarwal2023} are detailed in Appendix~\ref{sec:appendix-bert} and~\ref{sec:appendix-graphNLI} respectively, exhibiting performance trends comparable to those of \texttt{RoBERTa-Large} and hierarchical models.

Baseline models were trained according to their original implementations, with minor adjustments to accommodate different dataset sizes (e.g., increased epochs upon smaller training sets). Performance metrics represent the average across five random seeds, where training subsets are also independently sampled for each seed (for WIKI and CMV, conversation pairs are sampled accordingly). Training details are provided in Appendix~\ref{sec:appendix-experiment-detail}.

Other recent methods have been proposed, including summary-based~\cite{Github-Imran2025,Summary-Hua2024} and generation-based~\cite{LLMDerail-Zhang2025} approaches. They operate at conversation level, making a single prediction given the full dialogue or its summary, rather than at utterance-level for dynamic forecasting. The approach of~\citet{GraphCN-Altarawneh2023} relies on 
community voting scores on utterances, a feature available in CMV but absent in others. Some methods shift to dynamically training the 
models on incrementally expanding dialogue histories to improve performance or enable earlier detection~\cite{TransformerHierarchical-Yuan2023,dynamic-Kementchedjhieva2021}, which alters the training paradigm and is orthogonal to our modeling approach. Given our focus on applicability to different datasets, cross-dataset generalizability, and dynamic forecasting, we exclude these methods from our baseline comparisons. However, we note that their underlying architectures are primarily PLM-based and hierarchical, which are well-represented by our selected baselines.

\section{Results and Discussion}
\subsection{In-Domain Performance}\label{sec:in-domain-performance}

\begin{figure}[h]
    \centering
    \includegraphics[width=0.85\linewidth]{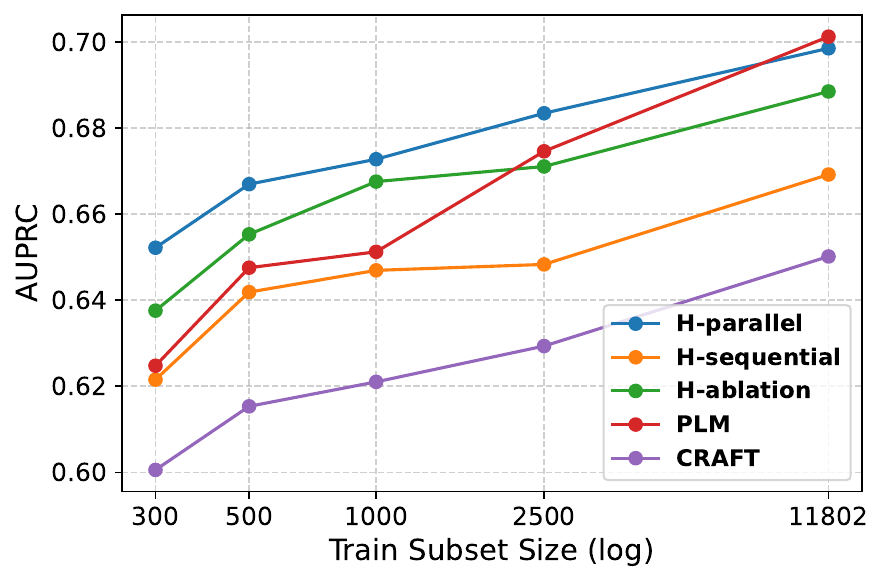}
    \caption{Models' $\text{Dynamic}_{\text{mean}}$ AUPRC learning curves on CMV, plotted against training subset size.}
    \label{fig:CMV-CMV-mean-AUPRC}
\end{figure}

Table~\ref{tab:in-domain-mean-AULC} reports $\text{Dynamic}_{\text{mean}}$ AULC, enabling comparisons between the models via single metrics. Figure~\ref{fig:CMV-CMV-mean-AUPRC} illustrates example learning curves across different training set sizes, while the AULC values reported in Table~\ref{tab:in-domain-mean-AULC} represent the area under these curves, scaled by the range of the x-axis to fall within [0,1].
From Table~\ref{tab:in-domain-mean-AULC}, $H_{\text{parallel}}$ consistently achieves the best performance across datasets\footnote{\label{fn:statistical-test}They achieve a statistically significant improvement ($p < 0.05$, paired t-test) over all baseline models on one or both of Macro-F1 and AUPRC, though 5 seeds has difficulty to test normality assumption and provides limited statistical power.}, with PLM and $H_{\text{ablation}}$ typically ranking second. 

All models except CRAFT are based on the \texttt{RoBERTa-Large} PLM. The latter is pre-trained on a large generic corpus, whereas CRAFT is pre-trained on a substantially smaller, domain-specific corpus\footnote{\label{fn:CRAFT}CRAFT is unavailable for GITHUB due to missing pretraining corpus.}. Despite this, CRAFT still demonstrates comparable performance as a lightweight model. 
The architectural shift from a flat PLM to a hierarchical design ($H_{\text{ablation}}$, which uses the PLM as an utterance encoder) leads to slightly improved AUPRC, though it can underperform on Macro-F1 in some cases.
The additional gains observed in $H_{\text{parallel}}$ over $H_{\text{ablation}}$ highlight the benefit of incorporating SA to guide model predictions.
In contrast, $H_{\text{sequential}}$ performs worse than most PLM-based models and shows similar performance to CRAFT. We hypothesise that this is because $H_{\text{sequential}}$ more strongly enforces reliance on SA knowledge compared to $H_{\text{parallel}}$, which can suppress the availability of generalizable textual features and lead to poorer performance in in-domain settings.

While $H_{\text{parallel}}$ consistently outperforms other models in $\text{Dynamic}_{\text{mean}}$, it does not in $\text{Dynamic}_{\text{max}}$ (reported in Appendix~\ref{sec:appendix-max}), instead ranking first or second across datasets. This indicates that the model achieves better performance when predictions are updated at each timestep, rather than fixed upon any derailment alarm. To further examine models' dynamic forecasting capacity, a per-timestep analysis is presented later in this section.

\paragraph{Model Performance Across Data Regimes}

\begin{figure*}[t]
    \centering

    \begin{subfigure}[b]{0.329\textwidth}
        \centering
        \includegraphics[width=1.03\textwidth]{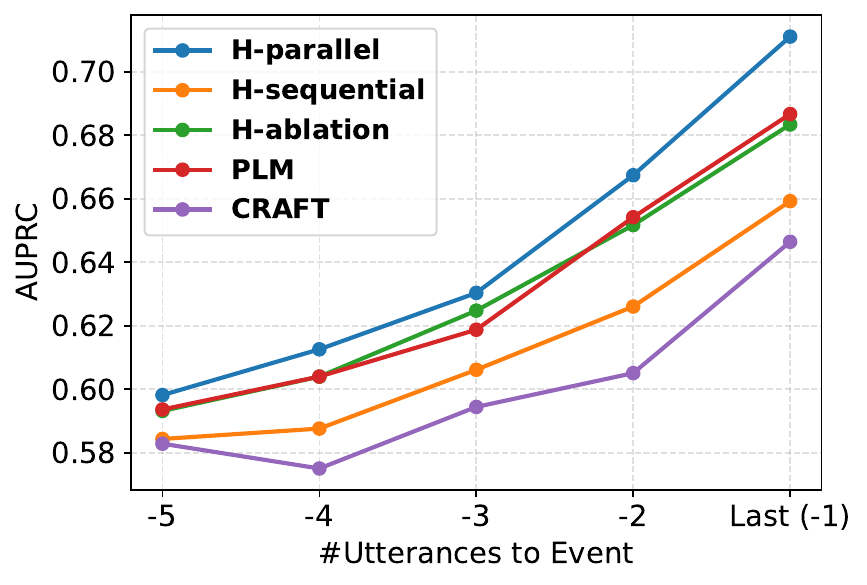}
        \caption{300 Train Subset}
    \end{subfigure}
    \hfill
    \begin{subfigure}[b]{0.329\textwidth}
        \centering
        \includegraphics[width=1.03\textwidth]{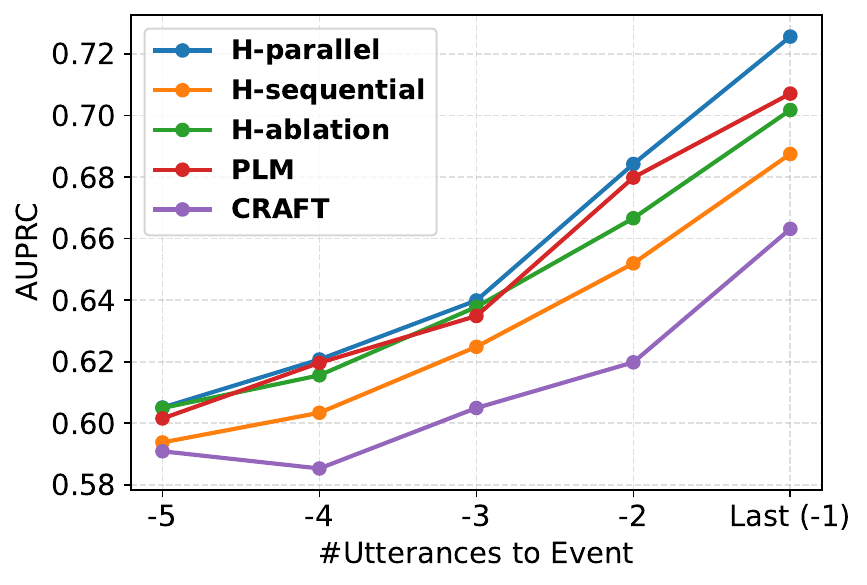}
        \caption{500 Train Subset}
    \end{subfigure}
    \hfill
    \begin{subfigure}[b]{0.329\textwidth}
        \centering
        \includegraphics[width=1.03\textwidth]{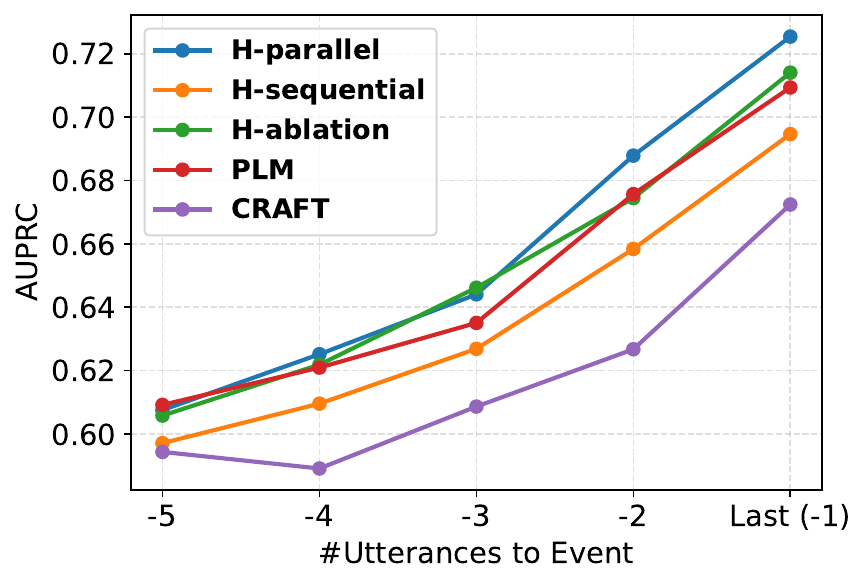}
        \caption{1000 Train Subset}
    \end{subfigure}

    \caption{Model AUPRC performance on CMV is plotted as dialogues progress toward derailment or conversation end, across training data sizes. 
    $H_{\text{parallel}}$ achieves the best performance, particularly in the low-data regime.}
    \label{fig:timestep-performance-CMV}
\end{figure*}

Figure~\ref{fig:CMV-CMV-mean-AUPRC} presents the $\text{Dynamic}_{\text{mean}}$ learning curve of the models prior to AULC aggregation. The CMV performance is visualised here; similar trends for other datasets are reported in Appendix~\ref{sec:appendix-learning-curve}. Notably, $H_{\text{parallel}}$ demonstrates strong performance in low-data regimes (300-2500 samples), a critical advantage for real-world scenarios with high annotation costs (e.g., GITHUB and WIKI datasets have approximately 500 and 2500 training samples respectively). As training size increases to 11802, the standard PLM marginally outperforms $H_{\text{parallel}}$. This indicates that pragmatic information provides effective guidance when training resources are limited. However, as the amount of training data increases, text-based model is capable of learning generalizable features from text alone, reducing the benefit of explicit SA knowledge guidance.

\paragraph{Per-Timestep Model Performance}
Figure~\ref{fig:timestep-performance-CMV} presents models' per-timestep performance as dialogues progress toward derailment or conversation end, avoiding temporal aggregation via either $\text{Dynamic}_{\text{max}}$ or $\text{Dynamic}_{\text{mean}}$. We report CMV performance for up to 1000 samples here; results for the full dataset and other datasets show similar trends and are provided in Appendix~\ref{sec:appendix-timestep-performance}.
All models show improving performance as dialogues approach the target event, as derailment signals in the conversation build up and become more apparent. 
The relative ranking of the lines is largely consistent with the model rankings observed in Figure~\ref{fig:CMV-CMV-mean-AUPRC} for $\text{Dynamic}_{\text{mean}}$ performance. 
$H_{\text{parallel}}$ achieves the highest performance across individual timesteps, particularly in the low-data regime, with the gap between $H_{\text{parallel}}$ and other models closing as training subset size increases. These results align with our previous discussion, demonstrating that $H_{\text{parallel}}$ provides better forecasting throughout conversations, especially under data constraints.

\subsection{Cross-Dataset Generalization}

\begin{table*}[t]
\centering
\small
\setlength{\tabcolsep}{4pt}
\begin{tabular}{lcccccccccccc}
\toprule
& \multicolumn{2}{c}{WIKI$\rightarrow$CMV}
& \multicolumn{2}{c}{WIKI$\rightarrow$GITHUB}
& \multicolumn{2}{c}{CMV$\rightarrow$WIKI}
& \multicolumn{2}{c}{CMV$\rightarrow$GITHUB}
& \multicolumn{2}{c}{GITHUB$\rightarrow$WIKI}
& \multicolumn{2}{c}{GITHUB$\rightarrow$CMV} \\
\cmidrule(lr){2-3} \cmidrule(lr){4-5} \cmidrule(lr){6-7}
\cmidrule(lr){8-9} \cmidrule(lr){10-11} \cmidrule(lr){12-13}

Model
& F1 & AUPR
& F1 & AUPR
& F1 & AUPR
& F1 & AUPR
& F1 & AUPR
& F1 & AUPR \\
\midrule
Random
& $50.00$ & $50.00$
& $50.00$ & $23.33$
& $50.00$ & $50.00$
& $50.00$ & $23.33$
& $50.00$ & $50.00$
& $50.00$ & $50.00$\\
\hline
CRAFT\textsuperscript{\ref{fn:CRAFT}} 
& $44.62^\dagger$ & $57.45^\dagger$
& $61.02^\dagger$ & $44.65^\dagger$
& $53.31^\dagger$ & $62.14^\dagger$
& $51.75^\dagger$ & $40.09^\dagger$
& -- & --
& -- & -- \\

PLM
& \underline{$49.10$} & \underline{$61.08^\dagger$}
& $59.07^\dagger$ & $41.55^\dagger$
& \underline{$60.99$} & $65.88^\dagger$
& $58.75^\dagger$ & $42.93^\dagger$
& $46.00^\dagger$ & $59.14^\dagger$
& $34.52$ & \underline{$56.28^\dagger$} \\

$H_{\text{ablation}}$
& $47.24$ & $59.70^\dagger$
& $57.52^\dagger$ & $41.22^\dagger$
& $60.00^\dagger$ & \underline{$66.36^\dagger$}
& $57.73^\dagger$ & $41.93^\dagger$
& $50.25$ & \underline{$61.45^\dagger$}
& $\mathbf{36.19}$ & $53.85^\dagger$ \\

\hline

$H_{\text{sequential}}$
& $46.90^\dagger$ & $59.93^\dagger$
& $\mathbf{65.49}$ & $\mathbf{56.63}$
& $59.84^\dagger$ & $65.33^\dagger$
& $\mathbf{64.23}$ & $\mathbf{53.06}$
& $\mathbf{56.66}$ & $60.20^\dagger$
& \underline{$34.92$} & $56.02^\dagger$ \\

$H_{\text{parallel}}$
& $\mathbf{51.19}$ & $\mathbf{62.53}$
& \underline{$62.94$} & \underline{$54.15$}
& $\mathbf{61.49}$ & $\mathbf{67.14}$
& \underline{$60.38^\dagger$} & \underline{$48.42^\dagger$}
& \underline{$50.67^\dagger$} & $\mathbf{66.46}$
& $34.19$ & $\mathbf{57.63}$ \\

\bottomrule
\end{tabular}
\caption{Cross-dataset evaluation of $\text{Dynamic}_{\text{mean}}$ AULC using Macro-F1 (F1) and AUPRC (AUPR), standard deviations are omitted for brevity. Each configuration $D1\rightarrow D2$ denotes training on $D1$ and testing on $D2$. Best results are \textbf{bolded} and second-best are \underline{underlined}. The $\dagger$ symbol denotes that the model's performance is statistically significantly different to the best model ($p < 0.05$, two-tailed paired t-test). SA-informed models ($H_{\text{sequential}}$ and $H_{\text{parallel}}$) achieve strong cross-dataset performance.}
\label{tab:cross-domain-mean-AULC}
\end{table*}

Table~\ref{tab:cross-domain-mean-AULC} reports cross-dataset $\text{Dynamic}_{\text{mean}}$ AULC performance, evaluating the models' capacity to generalize to different domains. 

Overall, the SA-informed models ($H_{\text{parallel}}$ and $H_{\text{sequential}}$) yield superior performance\textsuperscript{\ref{fn:statistical-test}}. Experiments integrating SA information into GraphNLI~\cite{GraphNLI-Agarwal2023} (Appendix~\ref{sec:appendix-graphNLI}) also demonstrate consistent in-domain and cross-dataset performance improvements. This demonstrates that explicit SA modeling provides transferable signals for derailment forecasting upon domain shifts. 
An exception occurs in the GITHUB $\rightarrow$ CMV transfer, where $H_{\text{ablation}}$ achieves the highest Macro-F1, while $H_{\text{parallel}}$ achieves the highest AUPRC. Transferring from a small and technical dataset (GITHUB) to an open-domain dataset (CMV) is inherently difficult, as reflected in uniformly low scores across all models (Macro-F1 peaking at only 36.19). Under such domain shift, both the learned signals and the source-tuned thresholds degrade. The discrepancy between the winning models in the two metrics likely stems from their respective evaluation mechanics: Macro-F1 is further penalized by the failure of a source-tuned threshold, whereas AUPRC is threshold-independent. While transferability remains limited in this setting, AUPRC offers a lens into the models' residual capacity to correctly rank derailment.

Among the baseline models, CRAFT exhibits the lowest transferability due to its smaller architecture and domain-specific pre-training, compared to the other models which rely on large PLMs.

\subsection{Dataset Deviation and Model Generalization} 
An analysis of dataset deviations provides further insight into model generalization: while SA information generally improves cross-dataset generalization, prioritizing SA features yields the greatest benefit when transferring to highly divergent datasets. We quantify cross-dataset semantic and lexical deviations using Wasserstein distance on embedding distributions~\cite{CorpusEvaluation-Kour2022, WassersteinDistance-Heusel2017} and CHI distance on token frequencies~\cite{CorpusEvaluation-Kour2022,CHIDistance-Kilgarriff2001}, respectively.
3000 randomly sampled responses from each dataset were encoded using a sentence transformer (\texttt{all-mpnet-base-v2}) for semantic measurement, and tokenized via \texttt{SpaCy} (\texttt{en\_core\_web\_sm}) for the lexical evaluation.

\begin{figure}[h]
    \centering

    \begin{subfigure}[b]{0.48\linewidth}
        \centering
        \includegraphics[width=\linewidth]{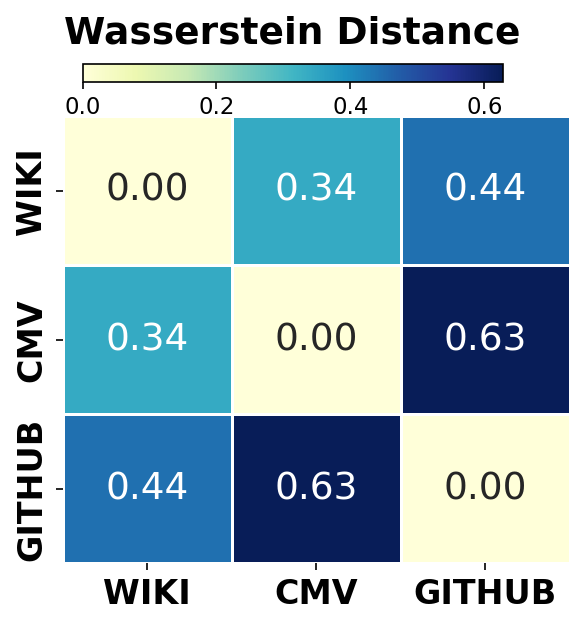}
        \caption{Semantic Deviation}
    \end{subfigure}
    \hfill
    \begin{subfigure}[b]{0.48\linewidth}
        \centering
        \includegraphics[width=\linewidth]{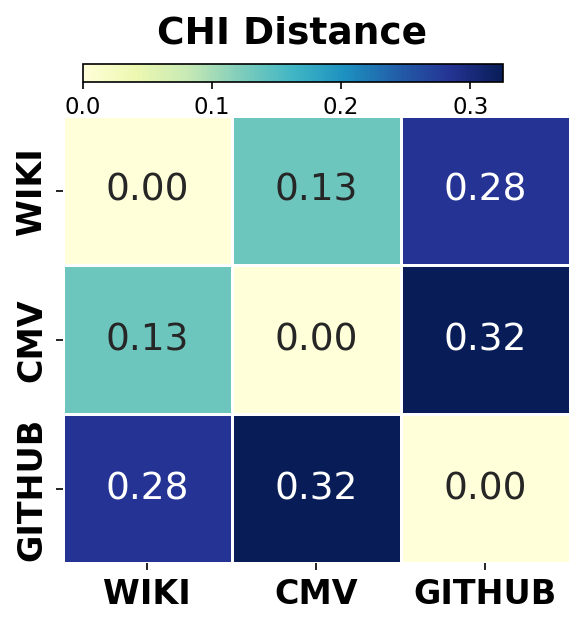}
        \caption{Lexical Deviation}
    \end{subfigure}
    \caption{Pairwise semantic Wasserstein distance and lexical CHI distance between datasets. Darker shades indicate higher deviation. GITHUB has the highest deviation and is more similar to WIKI than to CMV.}
    \label{fig:dataset-deviation}
\end{figure}

As shown in Figure~\ref{fig:dataset-deviation}, GITHUB exhibits the greatest overall deviation from the other two corpora, although it remains marginally more similar to WIKI than to CMV. This divergence aligns with the underlying nature of their conversational settings. Specifically, WIKI and CMV represent less specialised environments, encompassing diverse topics across editorial discussions and opinion debates, whereas GITHUB is highly domain-specific, focusing on technical problem-solving with extensive technical jargon and code snippets. The slight similarity between WIKI and GITHUB can be attributed to their shared task-oriented nature, in contrast to the debate-driven interactions of CMV.

When transferring from WIKI and CMV to the divergent GITHUB dataset, $H_{\text{sequential}}$ demonstrates the highest performance, followed by $H_{\text{parallel}}$. We attribute this to the fact that models relying on raw text can struggle with domain-specific overfitting, whereas the pragmatic SA representation prioritized by $H_{\text{sequential}}$ remain more robust to these lexical shifts, making them more effective for cross-domain generalization. 
Conversely, in transfers between the more lexically similar WIKI and CMV datasets, $H_{\text{parallel}}$ excels by combining both text and SA features, with text-based models (PLM and $H_{\text{ablation}}$) generally ranking second. 
Finally, training on the smaller, domain-specific GITHUB corpus yields uniformly lower transferability, with GITHUB $\rightarrow$ WIKI and GITHUB $\rightarrow$ CMV consistently underperform their CMV- and WIKI-trained counterparts. Yet, the AUPRC AULC achieved by $H_{\text{parallel}}$ when trained on the small GITHUB dataset (capped at 300-500 samples) and tested on WIKI (66.56\%) is highly comparable to WIKI's best in-domain performance (66.94\%, capped to 300-2500 samples). This underscores the strong cross-dataset transferability of our proposed approach.

\section{Conclusion}
In this work, we propose modeling speech acts as a pragmatic signal to improve derailment forecasting performance under low-data and cross-domain settings.\footnote{Code: \url{https://anonymous.4open.science/r/SA_Conversation_Derailment_Forecast-1242/}} We present two architectures that integrate SA information as an auxiliary learning signal alongside textual semantics. Extensive experiments demonstrate consistent improvements over strong baselines. This work contributes towards more robust dynamic moderation systems in diverse and data-scarce real-world settings.

\section*{Limitations}
We acknowledge certain limitations in our work. 
First, LLM-based SA extraction may introduce biases related to culture, dialect, and community norms, from their training data. The models are most appropriately used alongside human supervision in making proactive decisions, and not as fully autonomous systems.
Second, while our approach seeks to capture the intended meaning rather than literal meaning, LLMs are known to struggle to reliably recognize indirect intents, which is an active area of NLP research~\cite{Indirectness-Orsini2025}. This failure to capture true underlying intent can propagate sub-optimal signals downstream. Although our supplementary experiments show our framework is robust to SA granularity, this lack of reliance on highly nuanced intents may be an artifact of the LLM-based extraction. Specifically, the noise in LLM-derived labels may force the downstream model to treat SAs simply as a compact pragmatic representation, rather than learning from them to identify indirect, hidden intents. Future work could examine if introducing manually extracted, ground-truth SAs can provide the clean signal needed for the model to learn from these nuanced intents. 
Lastly, although our approach eliminates the need for SA extraction during inference, LLM-based or manual SA extraction still incurs time and computational costs during training. Future work could investigate more cost-effective alternatives, such as supervised SA extraction models, while maintaining accuracy. 

\section*{Acknowledgments}
This work was supported in part by the Australian Research Council Centre of Excellence for Automated Decision-Making and Society, and by the Australian Internet Observatory, which is co-funded by the Australian Research Data Commons (ARDC) through the HASS and Indigenous Research Data Commons.

\bibliography{custom}

@incollection{SAMeaning-Goddard2013,
    author = {Goddard, Cliff and Wierzbicka, Anna},
    isbn = {9780199668434},
    title = {Suggesting, apologizing, complimenting: English speech-act verbs},
    booktitle = {Words and Meanings: Lexical Semantics Across Domains, Languages, and Cultures},
    publisher = {Oxford University Press},
    year = {2013},
    month = {11},
    doi = {10.1093/acprof:oso/9780199668434.003.0007},
    url = {https://doi.org/10.1093/acprof:oso/9780199668434.003.0007},
    eprint = {https://academic.oup.com/book/0/chapter/148303263/chapter-pdf/38968736/acprof-9780199668434-chapter-7.pdf},
}

@inproceedings{SAParliamentaryDebates-Reinig2024,
    title = "How to Do Politics with Words: Investigating Speech Acts in Parliamentary Debates",
    author = "Reinig, Ines  and
      Rehbein, Ines  and
      Ponzetto, Simone Paolo",
    editor = "Calzolari, Nicoletta  and
      Kan, Min-Yen  and
      Hoste, Veronique  and
      Lenci, Alessandro  and
      Sakti, Sakriani  and
      Xue, Nianwen",
    booktitle = "Proceedings of the 2024 Joint International Conference on Computational Linguistics, Language Resources and Evaluation (LREC-COLING 2024)",
    month = may,
    year = "2024",
    address = "Torino, Italia",
    publisher = "ELRA and ICCL",
    url = "https://aclanthology.org/2024.lrec-main.727/",
    pages = "8287--8300"
}

@inproceedings{SATwitter-Saha2021,
    title = "Towards Sentiment and Emotion aided Multi-modal Speech Act Classification in {T}witter",
    author = "Saha, Tulika  and
      Upadhyaya, Apoorva  and
      Saha, Sriparna  and
      Bhattacharyya, Pushpak",
    editor = "Toutanova, Kristina  and
      Rumshisky, Anna  and
      Zettlemoyer, Luke  and
      Hakkani-Tur, Dilek  and
      Beltagy, Iz  and
      Bethard, Steven  and
      Cotterell, Ryan  and
      Chakraborty, Tanmoy  and
      Zhou, Yichao",
    booktitle = "Proceedings of the 2021 Conference of the North American Chapter of the Association for Computational Linguistics: Human Language Technologies",
    month = jun,
    year = "2021",
    address = "Online",
    publisher = "Association for Computational Linguistics",
    url = "https://aclanthology.org/2021.naacl-main.456/",
    doi = "10.18653/v1/2021.naacl-main.456",
    pages = "5727--5737"
}

@inproceedings{SAPoliticalCampaign-Subramanian2019,
    title = "Target Based Speech Act Classification in Political Campaign Text",
    author = "Subramanian, Shivashankar  and
      Cohn, Trevor  and
      Baldwin, Timothy",
    editor = "Mihalcea, Rada  and
      Shutova, Ekaterina  and
      Ku, Lun-Wei  and
      Evang, Kilian  and
      Poria, Soujanya",
    booktitle = "Proceedings of the Eighth Joint Conference on Lexical and Computational Semantics (*{SEM} 2019)",
    month = jun,
    year = "2019",
    address = "Minneapolis, Minnesota",
    publisher = "Association for Computational Linguistics",
    url = "https://aclanthology.org/S19-1030/",
    doi = "10.18653/v1/S19-1030",
    pages = "273--282"
}

@Article{SAOpenDomain-Compagno2018,
  author    = {Compagno, Dario and Epure, Elena V. and Deneckere-Lebas, Rébecca and Salinesi, Camille},
  journal   = {Corpus Pragmatics},
  title     = {Exploring Digital Conversation Corpora with Process Mining},
  year      = {2018},
  issn      = {2509-9515},
  month     = Feb,
  number    = {2},
  pages     = {193--215},
  volume    = {2},
  doi       = {10.1007/s41701-018-0030-6},
  publisher = {Springer Science and Business Media LLC},
}

@Book{SAMeaning-Vanderveken1990,
  author    = {Vanderveken, Daniel},
  publisher = {Cambridge University Press},
  title     = {Meaning and Speech Acts: Volume 1, Principles of Language Use},
  year      = {1990},
  address   = {Cambridge},
  isbn      = {9780521374156},
  volume    = {1},
}

@inbook{SATaxonomy-Searle1975, 
place={Cambridge}, 
title={A Taxonomy of Illocutionary Acts}, booktitle={Expression and Meaning: Studies in the Theory of Speech Acts}, 
publisher={Cambridge University Press}, author={Searle, John R.}, 
year={1979}, 
pages={1–29}
}

@Book{SAMeaning-Wierzbicka1987,
  author    = {Wierzbicka, Anna},
  publisher = {Academic Press},
  title     = {English speech act verbs: A Semantic Dictionary},
  year      = {1987},
  address   = {Sydney},
  isbn      = {9780123128102},
  pagetotal = {397},
}

@online{CambridgeDictionary,
  author = {{Cambridge University Press}},
  title = {Cambridge Dictionary},
  year = {2026},
  url = {https://dictionary.cambridge.org/},
  organization = {Cambridge Dictionary},
  urldate = {2026-05-11}
}

@article{SAUsage-Kholid2024,
author = {Kholid, Idham and Hidayat, Agus and Triyani, Bela and Stkip Pgri Bandar Lampung, Ijlhe},
year = {2024},
month = {12},
pages = {305-314},
title = {An Analysis of Assertive and Commissive Speech Acts in Simon Sinek's Speeches},
volume = {7},
journal = {IJLHE: International Journal of Language, Humanities, and Education},
doi = {10.52217/ijlhe.v7i2.1649}
}

@Article{SAUsage-Isnaeni2025,
  author       = {Isnaeni, Isnaeni and R., Mantasiah and Hasmawati, Hasmawati},
  journal      = {KLASIKAL : JOURNAL OF EDUCATION, LANGUAGE TEACHING AND SCIENCE},
  title        = {THE FUNCTION OF ASSERTIVE SPEECH ACTS IN THE NOVEL CALABAI BY PEPI AL-BAYQUNIE},
  year         = {2025},
  month        = {Apr.},
  number       = {1},
  pages        = {468–480},
  volume       = {7},
  doi          = {10.52208/klasikal.v7i1.1305},
  url          = {https://journalfkipuniversitasbosowa.org/index.php/klasikal/article/view/1305},
}

@inproceedings{DialogGLP-Wang2023,
    title = "Global-Local Modeling with Prompt-Based Knowledge Enhancement for Emotion Inference in Conversation",
    author = "Wang, Renxi  and
      Feng, Shi",
    editor = "Vlachos, Andreas  and
      Augenstein, Isabelle",
    booktitle = "Findings of the Association for Computational Linguistics: EACL 2023",
    month = may,
    year = "2023",
    address = "Dubrovnik, Croatia",
    publisher = "Association for Computational Linguistics",
    url = "https://aclanthology.org/2023.findings-eacl.158/",
    doi = "10.18653/v1/2023.findings-eacl.158",
    pages = "2120--2127"
}

@inproceedings{LLMDerail-Zhang2025,
    title = "Forecasting Conversation Derailments Through Generation",
    author = "Zhang, Yunfan  and
      McKeown, Kathleen  and
      Muresan, Smaranda",
    editor = "Flek, Lucie  and
      Narayan, Shashi  and
      Phương, L{\^e} Hồng  and
      Pei, Jiahuan",
    booktitle = "Proceedings of the 18th International Natural Language Generation Conference",
    month = oct,
    year = "2025",
    address = "Hanoi, Vietnam",
    publisher = "Association for Computational Linguistics",
    url = "https://aclanthology.org/2025.inlg-main.40/",
    pages = "699--715"
}

@inproceedings{IntentPerception-Chang2020,
author = {Chang, Jonathan P. and Cheng, Justin and Danescu-Niculescu-Mizil, Cristian},
title = {Don’t Let Me Be Misunderstood:Comparing Intentions and Perceptions in Online Discussions},
year = {2020},
isbn = {9781450370233},
publisher = {Association for Computing Machinery},
address = {New York, NY, USA},
url = {https://doi.org/10.1145/3366423.3380273},
doi = {10.1145/3366423.3380273},
booktitle = {Proceedings of The Web Conference 2020},
pages = {2066–2077},
numpages = {12},
location = {Taipei, Taiwan},
series = {WWW '20}
}

@misc{gptoss120b,
      title={gpt-oss-120b \& gpt-oss-20b Model Card}, 
      author={OpenAI},
      year={2025},
      eprint={2508.10925},
      archivePrefix={arXiv},
      primaryClass={cs.CL},
      url={https://arxiv.org/abs/2508.10925}, 
}

@inproceedings{RSA-spinosoDiPiano2025,
    title = "({RSA}){\texttwosuperior}: A Rhetorical-Strategy-Aware Rational Speech Act Framework for Figurative Language Understanding",
    author = "Spinoso-Di Piano, Cesare  and
      Austin, David Eric  and
      Piantanida, Pablo  and
      Cheung, Jackie CK",
    editor = "Che, Wanxiang  and
      Nabende, Joyce  and
      Shutova, Ekaterina  and
      Pilehvar, Mohammad Taher",
    booktitle = "Proceedings of the 63rd Annual Meeting of the Association for Computational Linguistics (Volume 1: Long Papers)",
    month = jul,
    year = "2025",
    address = "Vienna, Austria",
    publisher = "Association for Computational Linguistics",
    url = "https://aclanthology.org/2025.acl-long.1019/",
    doi = "10.18653/v1/2025.acl-long.1019",
    pages = "20898--20938",
    ISBN = "979-8-89176-251-0"
}

@inproceedings{Diplomat-Li2023,
 author = {Li, Hengli and Zhu, Song-Chun and Zheng, Zilong},
 booktitle = {Advances in Neural Information Processing Systems},
 editor = {A. Oh and T. Naumann and A. Globerson and K. Saenko and M. Hardt and S. Levine},
 pages = {46856--46884},
 publisher = {Curran Associates, Inc.},
 title = {Diplomat: A Dialogue Dataset for Situated PragMATic Reasoning},
 url = {https://proceedings.neurips.cc/paper_files/paper/2023/file/924303c6a45685510877ee018cdc8f80-Paper-Datasets_and_Benchmarks.pdf},
 volume = {36},
 year = {2023}
}

@inproceedings{CyberAggression-Ollagnier2024,
    title = "{C}yber{A}gression{A}do-v2: Leveraging Pragmatic-Level Information to Decipher Online Hate in {F}rench Multiparty Chats",
    author = "Ollagnier, Anais",
    editor = "Calzolari, Nicoletta  and
      Kan, Min-Yen  and
      Hoste, Veronique  and
      Lenci, Alessandro  and
      Sakti, Sakriani  and
      Xue, Nianwen",
    booktitle = "Proceedings of the 2024 Joint International Conference on Computational Linguistics, Language Resources and Evaluation (LREC-COLING 2024)",
    month = may,
    year = "2024",
    address = "Torino, Italia",
    publisher = "ELRA and ICCL",
    url = "https://aclanthology.org/2024.lrec-main.383/",
    pages = "4287--4298"
}

@inproceedings{Indirectness-Orsini2025,
    title = "Direct and Indirect Interpretations of Speech Acts: Evidence from Human Judgments and Large Language Models",
    author = "Orsini, Massimiliano  and
      Brunato, Dominique",
    editor = "Bosco, Cristina  and
      Jezek, Elisabetta  and
      Polignano, Marco  and
      Sanguinetti, Manuela",
    booktitle = "Proceedings of the Eleventh Italian Conference on Computational Linguistics (CLiC-it 2025)",
    month = sep,
    year = "2025",
    address = "Cagliari, Italy",
    publisher = "CEUR Workshop Proceedings",
    url = "https://aclanthology.org/2025.clicit-1.79/",
    pages = "837--848",
    ISBN = "979-12-243-0587-3"
}

@inproceedings{CRAFT-CMV-Chang2019,
    title = "Trouble on the Horizon: Forecasting the Derailment of Online Conversations as they Develop",
    author = "Chang, Jonathan P.  and
      Danescu-Niculescu-Mizil, Cristian",
    editor = "Inui, Kentaro  and
      Jiang, Jing  and
      Ng, Vincent  and
      Wan, Xiaojun",
    booktitle = "Proceedings of the 2019 Conference on Empirical Methods in Natural Language Processing and the 9th International Joint Conference on Natural Language Processing (EMNLP-IJCNLP)",
    month = nov,
    year = "2019",
    address = "Hong Kong, China",
    publisher = "Association for Computational Linguistics",
    url = "https://aclanthology.org/D19-1481/",
    doi = "10.18653/v1/D19-1481",
    pages = "4743--4754"
}

@misc{Evaluate-CMVLarge-Tran2025,
      title={Conversations Gone Awry, But Then? Evaluating Conversational Forecasting Models}, 
      author={Son Quoc Tran and Tushaar Gangavarapu and Nicholas Chernogor and Jonathan P. Chang and Cristian Danescu-Niculescu-Mizil},
      year={2025},
      eprint={2507.19470},
      archivePrefix={arXiv},
      primaryClass={cs.CL},
      url={https://arxiv.org/abs/2507.19470}, 
}

@inproceedings{Wiki-Zhang2018,
    title = "Conversations Gone Awry: Detecting Early Signs of Conversational Failure",
    author = "Zhang, Justine  and
      Chang, Jonathan  and
      Danescu-Niculescu-Mizil, Cristian  and
      Dixon, Lucas  and
      Hua, Yiqing  and
      Taraborelli, Dario  and
      Thain, Nithum",
    editor = "Gurevych, Iryna  and
      Miyao, Yusuke",
    booktitle = "Proceedings of the 56th Annual Meeting of the Association for Computational Linguistics (Volume 1: Long Papers)",
    month = jul,
    year = "2018",
    address = "Melbourne, Australia",
    publisher = "Association for Computational Linguistics",
    url = "https://aclanthology.org/P18-1125/",
    doi = "10.18653/v1/P18-1125",
    pages = "1350--1361"
}

@misc{Github-Imran2025,
      title={Understanding and Predicting Derailment in Toxic Conversations on GitHub}, 
      author={Mia Mohammad Imran and Robert Zita and Rebekah Copeland and Preetha Chatterjee and Rahat Rizvi Rahman and Kostadin Damevski},
      year={2025},
      eprint={2503.02191},
      archivePrefix={arXiv},
      primaryClass={cs.SE},
      url={https://arxiv.org/abs/2503.02191}, 
}

@article{AULC-Viering2023,
author = {Viering, Tom and Loog, Marco},
title = {The Shape of Learning Curves: A Review},
year = {2023},
issue_date = {June 2023},
publisher = {IEEE Computer Society},
address = {USA},
volume = {45},
number = {6},
issn = {0162-8828},
url = {https://doi.org/10.1109/TPAMI.2022.3220744},
doi = {10.1109/TPAMI.2022.3220744},
journal = {IEEE Trans. Pattern Anal. Mach. Intell.},
month = jun,
pages = {7799–7819},
numpages = {21}
}

@article{roberta-Liu2018,
  author    = {Yinhan Liu and
               Myle Ott and
               Naman Goyal and
               Jingfei Du and
               Mandar Joshi and
               Danqi Chen and
               Omer Levy and
               Mike Lewis and
               Luke Zettlemoyer and
               Veselin Stoyanov},
  title     = {RoBERTa: {A} Robustly Optimized {BERT} Pretraining Approach},
  journal   = {CoRR},
  volume    = {abs/1907.11692},
  year      = {2019},
  url       = {http://arxiv.org/abs/1907.11692},
  archivePrefix = {arXiv},
  eprint    = {1907.11692},
  bibsource = {dblp computer science bibliography, https://dblp.org}
}

@inproceedings{Summary-Hua2024,
    title = "How did we get here? Summarizing conversation dynamics",
    author = "Hua, Yilun  and
      Chernogor, Nicholas  and
      Gu, Yuzhe  and
      Jeong, Seoyeon  and
      Luo, Miranda  and
      Danescu-Niculescu-Mizil, Cristian",
    editor = "Duh, Kevin  and
      Gomez, Helena  and
      Bethard, Steven",
    booktitle = "Proceedings of the 2024 Conference of the North American Chapter of the Association for Computational Linguistics: Human Language Technologies (Volume 1: Long Papers)",
    month = jun,
    year = "2024",
    address = "Mexico City, Mexico",
    publisher = "Association for Computational Linguistics",
    url = "https://aclanthology.org/2024.naacl-long.414/",
    doi = "10.18653/v1/2024.naacl-long.414",
    pages = "7452--7477"
}

@inproceedings{GraphCN-Altarawneh2023,
    title = "Conversation Derailment Forecasting with Graph Convolutional Networks",
    author = "Altarawneh, Enas  and
      Agrawal, Ameeta  and
      Jenkin, Michael  and
      Papagelis, Manos",
    editor = {Chung, Yi-Ling  and
      R{\"o}ttger, Paul  and
      Nozza, Debora  and
      Talat, Zeerak  and
      Mostafazadeh Davani, Aida},
    booktitle = "The 7th Workshop on Online Abuse and Harms (WOAH)",
    month = jul,
    year = "2023",
    address = "Toronto, Canada",
    publisher = "Association for Computational Linguistics",
    url = "https://aclanthology.org/2023.woah-1.16/",
    doi = "10.18653/v1/2023.woah-1.16",
    pages = "160--169"
}

@article{TransformerHierarchical-Yuan2023, 
title={Conversation Modeling to Predict Derailment}, 
volume={17}, 
url={https://ojs.aaai.org/index.php/ICWSM/article/view/22200}, 
DOI={10.1609/icwsm.v17i1.22200}, 
abstractNote={Conversations among online users sometimes derail, i.e., break down into personal attacks. Derailment interferes with the healthy growth of communities in cyberspace. The ability to predict whether an ongoing conversation will derail could provide valuable advance, even real-time, insight to both interlocutors and moderators. Prior approaches predict conversation derailment retrospectively without the ability to forestall the derailment proactively. Some existing works attempt to make dynamic predictions as the conversation develops, but fail to incorporate multisource information, such as conversational structure and distance to derailment. We propose a hierarchical transformer-based framework that combines utterance-level and conversation-level information to capture fine-grained contextual semantics. We propose a domain-adaptive pretraining objective to unite conversational structure information and a multitask learning scheme to leverage the distance from each utterance to derailment. An evaluation of our framework on two conversation derailment datasets shows an improvement in F1 score for the prediction of derailment. These results demonstrate the effectiveness of incorporating multisource information for predicting the derailment of a conversation.}, 
number={1}, 
journal={Proceedings of the International AAAI Conference on Web and Social Media}, 
author={Yuan, Jiaqing and Singh, Munindar P.}, 
year={2023}, 
month={Jun.}, 
pages={926–935} }

@inproceedings{dynamic-Kementchedjhieva2021,
    title = "Dynamic Forecasting of Conversation Derailment",
    author = "Kementchedjhieva, Yova  and
      S{\o}gaard, Anders",
    editor = "Moens, Marie-Francine  and
      Huang, Xuanjing  and
      Specia, Lucia  and
      Yih, Scott Wen-tau",
    booktitle = "Proceedings of the 2021 Conference on Empirical Methods in Natural Language Processing",
    month = nov,
    year = "2021",
    address = "Online and Punta Cana, Dominican Republic",
    publisher = "Association for Computational Linguistics",
    url = "https://aclanthology.org/2021.emnlp-main.624/",
    doi = "10.18653/v1/2021.emnlp-main.624",
    pages = "7915--7919"
}

@Article{EmotionACEIC-Xu2025,
  author    = {Xu, Xingle and Feng, Shi and Cui, Yuan and Zhang, Yifei and Wang, Daling},
  journal   = {International Journal of Machine Learning and Cybernetics},
  title     = {AC-EIC: addressee-centered emotion inference in conversations},
  year      = {2025},
  issn      = {1868-808X},
  month     = Feb,
  number    = {7-8},
  pages     = {5113--5130},
  volume    = {16},
  doi       = {10.1007/s13042-025-02561-9},
  publisher = {Springer Science and Business Media LLC},
}

@inproceedings{EmotionDialogueGLP-Wang2023,
    title = "Global-Local Modeling with Prompt-Based Knowledge Enhancement for Emotion Inference in Conversation",
    author = "Wang, Renxi  and
      Feng, Shi",
    editor = "Vlachos, Andreas  and
      Augenstein, Isabelle",
    booktitle = "Findings of the Association for Computational Linguistics: EACL 2023",
    month = may,
    year = "2023",
    address = "Dubrovnik, Croatia",
    publisher = "Association for Computational Linguistics",
    url = "https://aclanthology.org/2023.findings-eacl.158/",
    doi = "10.18653/v1/2023.findings-eacl.158",
    pages = "2120--2127"
}

@article{BERT-Devlin2018,
  author    = {Jacob Devlin and
               Ming{-}Wei Chang and
               Kenton Lee and
               Kristina Toutanova},
  title     = {{BERT:} Pre-training of Deep Bidirectional Transformers for Language
               Understanding},
  journal   = {CoRR},
  volume    = {abs/1810.04805},
  year      = {2018},
  url       = {http://arxiv.org/abs/1810.04805},
  archivePrefix = {arXiv},
  eprint    = {1810.04805},
  bibsource = {dblp computer science bibliography, https://dblp.org}
}

@inproceedings{CorpusEvaluation-Kour2022,
    title = "Measuring the Measuring Tools: An Automatic Evaluation of Semantic Metrics for Text Corpora",
    author = "Kour, George  and
      Ackerman, Samuel  and
      Farchi, Eitan Daniel  and
      Raz, Orna  and
      Carmeli, Boaz  and
      Tavor, Ateret Anaby",
    editor = "Bosselut, Antoine  and
      Chandu, Khyathi  and
      Dhole, Kaustubh  and
      Gangal, Varun  and
      Gehrmann, Sebastian  and
      Jernite, Yacine  and
      Novikova, Jekaterina  and
      Perez-Beltrachini, Laura",
    booktitle = "Proceedings of the Second Workshop on Natural Language Generation, Evaluation, and Metrics (GEM)",
    month = dec,
    year = "2022",
    address = "Abu Dhabi, United Arab Emirates (Hybrid)",
    publisher = "Association for Computational Linguistics",
    url = "https://aclanthology.org/2022.gem-1.35/",
    doi = "10.18653/v1/2022.gem-1.35",
    pages = "405--416"
}

@inproceedings{WassersteinDistance-Heusel2017,
author = {Heusel, Martin and Ramsauer, Hubert and Unterthiner, Thomas and Nessler, Bernhard and Hochreiter, Sepp},
title = {GANs trained by a two time-scale update rule converge to a local nash equilibrium},
year = {2017},
isbn = {9781510860964},
publisher = {Curran Associates Inc.},
address = {Red Hook, NY, USA},
booktitle = {Proceedings of the 31st International Conference on Neural Information Processing Systems},
pages = {6629–6640},
numpages = {12},
location = {Long Beach, California, USA},
series = {NIPS'17}
}

@Article{CHIDistance-Kilgarriff2001,
  author    = {Kilgarriff, Adam},
  journal   = {International Journal of Corpus Linguistics},
  title     = {Comparing Corpora},
  year      = {2001},
  issn      = {1569-9811},
  month     = Dec,
  number    = {1},
  pages     = {97--133},
  volume    = {6},
  doi       = {https://doi.org/10.1075/ijcl.6.1.05kil},
  publisher = {John Benjamins Publishing Company},
}

@Book{SpeechActTheory-Austin1975,
  author    = {John Langshaw Austin},
  publisher = {Oxford University Press},
  title     = {How To Do Things With Words},
  year      = {1975},
  isbn      = {9780198245537},
  month     = sep,
  doi       = {10.1093/acprof:oso/9780198245537.001.0001},
}

@inproceedings{SAGeneralizability-Aljanaideh2025,
    title = "Speech Act Patterns for Improving Generalizability of Explainable Politeness Detection Models",
    author = "Aljanaideh, Ahmad",
    editor = "Che, Wanxiang  and
      Nabende, Joyce  and
      Shutova, Ekaterina  and
      Pilehvar, Mohammad Taher",
    booktitle = "Findings of the Association for Computational Linguistics: ACL 2025",
    month = jul,
    year = "2025",
    address = "Vienna, Austria",
    publisher = "Association for Computational Linguistics",
    url = "https://aclanthology.org/2025.findings-acl.970/",
    doi = "10.18653/v1/2025.findings-acl.970",
    pages = "18945--18954",
    ISBN = "979-8-89176-256-5"
}

@inproceedings{TOD-Wu2020,
    title = "{TOD}-{BERT}: Pre-trained Natural Language Understanding for Task-Oriented Dialogue",
    author = "Wu, Chien-Sheng  and
      Hoi, Steven C.H.  and
      Socher, Richard  and
      Xiong, Caiming",
    editor = "Webber, Bonnie  and
      Cohn, Trevor  and
      He, Yulan  and
      Liu, Yang",
    booktitle = "Proceedings of the 2020 Conference on Empirical Methods in Natural Language Processing (EMNLP)",
    month = nov,
    year = "2020",
    address = "Online",
    publisher = "Association for Computational Linguistics",
    url = "https://aclanthology.org/2020.emnlp-main.66/",
    doi = "10.18653/v1/2020.emnlp-main.66",
    pages = "917--929"
}

@Article{hateSpeechDetection-Poletto2020,
  author    = {Poletto, Fabio and Basile, Valerio and Sanguinetti, Manuela and Bosco, Cristina and Patti, Viviana},
  journal   = {Language Resources and Evaluation},
  title     = {Resources and benchmark corpora for hate speech detection: a systematic review},
  year      = {2020},
  issn      = {1574-0218},
  month     = Sept,
  number    = {2},
  pages     = {477--523},
  volume    = {55},
  doi       = {https://doi.org/10.1007/s10579-020-09502-8},
  publisher = {Springer Science and Business Media LLC},
}

@INPROCEEDINGS {ASL-Ridnik2021,
author = { Ridnik, Tal and Ben-Baruch, Emanuel and Zamir, Nadav and Noy, Asaf and Friedman, Itamar and Protter, Matan and Zelnik-Manor, Lihi },
booktitle = { 2021 IEEE/CVF International Conference on Computer Vision (ICCV) },
title = {{ Asymmetric Loss For Multi-Label Classification }},
year = {2021},
volume = {},
ISSN = {},
pages = {82-91},
doi = {10.1109/ICCV48922.2021.00015},
url = {https://doi.ieeecomputersociety.org/10.1109/ICCV48922.2021.00015},
publisher = {IEEE Computer Society},
address = {Los Alamitos, CA, USA},
month =Oct}

@inproceedings{GRU-Cho2014,
    title = "Learning Phrase Representations using {RNN} Encoder{--}Decoder for Statistical Machine Translation",
    author = {Cho, Kyunghyun  and
      van Merri{\"e}nboer, Bart  and
      Gulcehre, Caglar  and
      Bahdanau, Dzmitry  and
      Bougares, Fethi  and
      Schwenk, Holger  and
      Bengio, Yoshua},
    editor = "Moschitti, Alessandro  and
      Pang, Bo  and
      Daelemans, Walter",
    booktitle = "Proceedings of the 2014 Conference on Empirical Methods in Natural Language Processing ({EMNLP})",
    month = oct,
    year = "2014",
    address = "Doha, Qatar",
    publisher = "Association for Computational Linguistics",
    url = "https://aclanthology.org/D14-1179/",
    doi = "10.3115/v1/D14-1179",
    pages = "1724--1734"
}

@article{GraphNLI-Agarwal2023,
author = {Agarwal, Vibhor and Young, Anthony P. and Joglekar, Sagar and Sastry, Nishanth},
title = {A Graph-Based Context-Aware Model to Understand Online Conversations},
year = {2023},
issue_date = {February 2024},
publisher = {Association for Computing Machinery},
address = {New York, NY, USA},
volume = {18},
number = {1},
issn = {1559-1131},
url = {https://doi.org/10.1145/3624579},
doi = {10.1145/3624579},
journal = {ACM Trans. Web},
month = nov,
articleno = {10},
numpages = {27}
}

@article{DistilBERT-Sanh2019,
  title={DistilBERT, a distilled version of BERT: smaller, faster, cheaper and lighter},
  author={Victor Sanh and Lysandre Debut and Julien Chaumond and Thomas Wolf},
  journal={ArXiv},
  year={2019},
  volume={abs/1910.01108}
}

\appendix

\section{Speech Act Extraction}
\label{sec:appendix-SA-extraction}

\subsection{Speech Act Taxonomy}
Various speech act taxonomies have been developed to model conversational dynamics in specialized domains, such as parliamentary debates~\cite{SAParliamentaryDebates-Reinig2024}, political campaign~\cite{SAPoliticalCampaign-Subramanian2019}, and Twitter text~\cite{SATwitter-Saha2021}. They typically derive SA labels from frequent communicative intents observed within their specific contexts.
However, conversational derailment can occur across diverse argumentative and collaborative settings, necessitating a more open-domain SA taxonomy.

The first effort toward an open-domain taxonomy for digital corpora was conducted by~\cite{SAOpenDomain-Compagno2018}. They build on the SA verbs defined in~\cite{SAMeaning-Vanderveken1990}, which are structured into five hierarchical trees corresponding to Searle’s five fundamental SA categories~\cite{SATaxonomy-Searle1975}. By selecting the 21 immediate children of the root nodes and introducing organizational layers with oppositional traits (e.g., context-dependency, sentiment and strength), they refined the set into 17 distinct speech acts.

Despite this foundation, restricting the taxonomy to the immediate children of the root nodes introduces semantic gaps and do not meet the granularity commonly observed in pragmatic analysis works. In the hierarchical trees, a successor’s illocutionary force is derived by adding components or increasing the strength of its parent’s force. Consequently, lower-level verbs often carry important nuances that are not fully captured by higher-level verbs (e.g., ``questioning'' expressing doubt compared to the neutral information-seeking of ``inquiring''). Furthermore, several relevant verbs (e.g., ``boasting'') were defined but omitted from the trees. 

To address these gaps, we adapted and expanded the existing taxonomy to cover commonly analyzed conversational speech acts. The expanded taxonomy preserves the structure of the original, allowing the extracted SAs to be merged and collapsed into the lower-grained taxonomy. Extracting SAs at a high granularity thus provide the flexibility to evaluate downstream performance at different levels of granularity. Given that comprehensive list exceeding 300 verbs are computationally infeasible for use as a label set, we adopted a multi-stage refinement process: 
\begin{enumerate}
    \item Sourcing: Beyond the original taxonomy, we aggregated frequently utilized speech act verbs from linguistic and conversational analysis literature~\cite{SAMeaning-Goddard2013, SAUsage-Isnaeni2025, SAUsage-Kholid2024, SATaxonomy-Searle1975}.
    \item Defining: Each speech act is accompanied by a brief description of its meaning and, where applicable, the speaker’s psychological state, derived from semantic linguistic works and standard lexicons~\cite{SAMeaning-Vanderveken1990, SAMeaning-Goddard2013, SAMeaning-Wierzbicka1987, CambridgeDictionary}.
    \item Filtering: We reduced the set by merging verbs with largely overlapping semantic definitions to eliminate redundancy and labeling ambiguity, and based on empirical observations of SA frequency during the initial extraction phase.
\end{enumerate}

This process yielded 50 speech act verbs and an additional ``other'' label, detailed in Tables \ref{tab:speech-acts-1} and~\ref{tab:speech-acts-2}, and structured in \Cref{fig:speech-act-assertive,fig:speech-act-directive,fig:speech-act-commissive,fig:speech-act-expressive} following the organization of prior work~\cite{SAOpenDomain-Compagno2018}. The speech act taxonomy is used during speech act extraction, via the prompt structure shown in Table~\ref{tab:sa-extract-prompt}. While speech act extraction is performed at sentence level, responses were segmented into sentences via \texttt{SpaCy} (\texttt{en\_core\_web\_sm}).

\subsection{Speech Act Extraction Alignment}

We follow prior work~\cite{SAOpenDomain-Compagno2018} to measure SA extraction agreement, replacing Fleiss’ Kappa with Cohen's Kappa, as Fleiss’ Kappa is designed for multiple-rater settings where Cohen's Kappa is designed for two-rater settings. For this evaluation, one of the authors manually labeled 85 sentences from 30 randomly selected responses (10 from each dataset), utilizing the established SA taxonomy and the LLM extraction prompt as reference guidelines. Table~\ref{tab:sa-agreement} reports the Cohen's Kappa scores between the manually extracted and LLM extracted SAs, along with the $z$-scores and $p$-values rejecting the null hypothesis of no agreement ($p < 0.01$ across all tests). The LLM demonstrated moderate agreement ($\kappa = 0.53$) across the 51 granular classes, and almost perfect agreement ($\kappa = 0.88$) when collapsed into the 5 fundamental categories (assertive, directive, commissive, expressive, and an `other' class replacing declarative). This aligns closely with the human inter-annotator agreement reported in prior literature~\cite{SAOpenDomain-Compagno2018}, suggesting that the LLM extractions could provide meaningful and human-interpretable pragmatic signals.

\begin{table}[h]
\centering
\begin{tabular}{lcc}
\toprule
 & \textbf{51 classes} & \textbf{5 categories} \\
\midrule
\textbf{Cohen's Kappa} & 0.53 & 0.88 \\
\textbf{z-test} & 26.60 & 8.79 \\
\textbf{p-value} & $<0.01$ & $<0.01$ \\
\textbf{Interpretation} & Moderate & Almost Perfect \\
\bottomrule
\end{tabular}
\caption{Cohen's Kappa agreement results between LLM-extracted and manually labeled SAs. The LLM demonstrates moderate and almost perfect agreement across the 51 fine-grained classes and the 5 fundamental categories, respectively.}
\label{tab:sa-agreement}
\end{table}

\section{Experiment Details}\label{sec:appendix-experiment-detail}

\subsection{Hyperparameters and Setup}
All settings for the CRAFT baseline follow its original implementation~\cite{CRAFT-CMV-Chang2019}, with the exception of an increased maximum number of epochs to accommodate smaller training datasets. For other architectures, we generally adopt the effective hyperparameters and setup identified in prior work for PLMs~\cite{Evaluate-CMVLarge-Tran2025, dynamic-Kementchedjhieva2021}. For fair comparison where the hierarchical models have access to anonymized speaker ID (e.g., Speaker 1, Speaker 2), the strings are also prepended to each utterance as input to PLM baselines. The maximum sequence length is capped at 512 tokens per dialogue (representing the maximum capacity for \texttt{RoBERTa-large} and \texttt{BERT}) and 128 tokens per utterance for hierarchical models, utilizing left truncation. The training batch size is set to 4 for \texttt{BERT} and 12 for \texttt{RoBERTa-large}. The learning rate is $5e-6$ for pre-trained parameters and $5e-5$ for newly initialized parameters. All models are trained with early stopping on the validation set. Table~\ref{tab:epoch} reports the maximum number of training epochs for each model under different sample sizes, determined by validation convergence.

\subsection{Model Implementation and Information}
We use the original implementation of CRAFT, with approximately 61M parameters. Other models were implemented using PyTorch (2.10.0) and the Transformers (4.57.2) library. All models were trained on a single 80GB NVIDIA A100 GPU. The \texttt{RoBERTa-large} baseline contains approximately 355M parameters. When utilizing \texttt{RoBERTa-large} as an utterance-level encoder, $H_{\text{parallel}}$ and $H_{\text{ablation}}$ have approximately 372M parameters, while $H_{\text{sequential}}$ has approximately 373M parameters. To provide a reference for computational requirements, a single training run on the complete CMV dataset requires approximately 3 hours for $H_{\text{parallel}}$, 2 hours for $H_{\text{sequential}}$ and $H_{\text{ablation}}$, 1 hour for \texttt{RoBERTa-large}, and under 10 minutes for CRAFT. We estimate the total computational budget for our main experimental training and evaluation to be approximately 300 GPU hours.

The SA extraction via \texttt{gpt-oss-120b}, with high reasoning effort and parallelized across 2 80GB NVIDIA A100 GPU, requires in average approximately 7 seconds per response, totaling approximately 180 GPU hours. This underscores the latency bottleneck inherent to LLM generation during real-time inference, motivating the use of LLM auxiliary information only during training to support high-traffic moderation systems.

\subsection{Optimization Objectives}
For the primary derailment forecasting task, we use Cross-Entropy loss. For SA detection task, we employ Binary Cross-Entropy (BCE) loss in $H_{\text{sequential}}$, a standard objective for multi-label classification. For $H_{\text{parallel}}$, we utilize Asymmetric Loss (ASL)~\cite{ASL-Ridnik2021}. ASL targets the severe positive-negative imbalance inherent to multi-label settings, where negative samples overwhelmingly dominate. It addresses this by down-weighting and hard-thresholding easy negatives during training (e.g., if the model predicts a negative probability below 5\%, the sample contributes zero loss, and the model stops attempting to push it closer to 0). However, this relaxed optimization can negatively impact $H_{\text{sequential}}$. Because $H_{\text{sequential}}$ relies exclusively on the detected SA probabilities to forecast derailment, these small, un-zeroed negative probabilities could introduce noise and create a pathway for the model to leak more raw textual semantics into the final layer. 

\subsection{Training Curriculum}
$H_{\text{parallel}}$ is trained using a two-phase approach to ensure the model effectively incorporates SA knowledge. During the first intermediate phase, the objective is solely SA detection. To prevent catastrophic forgetting and ensure the model does not overfit before learning the primary task, the bottom half of the PLM layers (12 layers for \texttt{RoBERTa-large}, 6 for \texttt{BERT}) are frozen, and the model is trained on a 60\%/40\% train-validation split of the training data. The second fine-tuning phase unfreezes all layers and employs a multi-task learning objective, weighting derailment forecasting and SA detection at 1.0 and 0.1, respectively, to maintain focus on the primary task. 
In contrast, because $H_{\text{sequential}}$ relies on the detected SA probabilities for forecasting, it structurally forces the model to primarily utilize SA knowledge. Therefore, it does not require an intermediate training phase and instead optimizes both tasks simultaneously with equal weights of 0.5. 
All models, aside from $H_{\text{parallel}}$, have all layers fully unfrozen throughout their entire fine-tuning process.

\begin{table*}[t]
\centering
\label{tab:epoch}
\begin{tabular}{lccccc}
\toprule
\textbf{\# Training Samples} & \textbf{300} & \textbf{500} & \textbf{1000} & \textbf{2500} & \textbf{11802} \\
\midrule
CRAFT      & 100 & 100 & 100 & 100 & 30 \\
PLM        & 20 & 20 & 10 & 10 & 4 \\
$H_{\text{ablation}}$   & 20 & 20 & 10 & 10 & 4 \\
$H_{\text{sequential}}$ & 50 & 30 & 20 & 10 & 4 \\
$H_{\text{parallel}}$ Intermediate   & 10/15\footnotemark[4] & 10 & 5  & 5  & 2 \\
$H_{\text{parallel}}$ Finetune   & 10 & 10 & 5  & 5  & 4 \\
\bottomrule
\end{tabular}
\caption{Number of training epochs used under different training set sizes.}
\label{tab:epoch}
\end{table*}

\subsection{Dataset}
All datasets utilized in this study (described in Section~\ref{sec:dataset}) are publicly available and distributed under the MIT License, which permits both academic and commercial use. These predominantly English datasets contain occurrences of offensive language, which were retained as they are essential for the derailment forecasting task. We did not collect any external personal data. While the raw datasets contain usernames, we utilized these solely to establish conversational structure (i.e., speaker turns) and replaced them with anonymous Speaker IDs (e.g. Speaker 1, Speaker 2) during model training to protect user privacy.

\section{Additional Experimental Results}
\subsection{$\text{Dynamic}_{\text{max}}$ Performance}\label{sec:appendix-max}
Table~\ref{tab:in-domain-max-AULC} reports the models' $\text{Dynamic}_{\text{max}}$ AULC performance. PLM and $H_{\text{parallel}}$ consistently rank first or second, suggesting they perform comparably well when dialogues are flagged as ``derailed'' upon ``derailed'' forecast at any timestep. 

Table~\ref{tab:cross-domain-max-AULC} reports the models' cross-dataset $\text{Dynamic}_{\text{max}}$ performance. Although the SA-informed models ($H_{\text{parallel}}$ and $H_{\text{sequential}}$) do not consistently achieve the best in-domain $\text{Dynamic}_{\text{max}}$ (from Table~\ref{tab:in-domain-max-AULC}), they demonstrate strong cross-dataset performance, achieving the highest rank in the majority of cases, while the PLM ranks first in two out of twelve instances.

While models are optimized to minimize forecasting error independently given a dialogue history, future work could explore optimizing directly for $\text{Dynamic}_{\text{max}}$, where a single threshold-crossing prediction dictates the classification of the entire sequence. This may better support moderation workflows that prioritize a single warning proactive intervention over continuous state-tracking.

\begin{table*}[t]
\centering
\small
\setlength{\tabcolsep}{6pt}
\begin{tabular}{lcccccc}
\toprule
& \multicolumn{2}{c}{WIKI} & \multicolumn{2}{c}{CMV} & \multicolumn{2}{c}{GITHUB} \\
\cmidrule(lr){2-3} \cmidrule(lr){4-5} \cmidrule(lr){6-7}
Model 
& MacroF1 & AUPRC
& MacroF1 & AUPRC
& MacroF1 & AUPRC \\
\midrule
Random
& $50.00$ & $50.00$
& $50.00$ & $50.00$
& $50.00$ & $23.33$\\
\hline
CRAFT\textsuperscript{\ref{fn:CRAFT}} 
& $61.56 \pm 0.40$ & $63.22 \pm 0.62^\dagger$
& $62.93 \pm 0.42^\dagger$ & $64.77 \pm 0.41^\dagger$
& -- & -- \\

PLM
& $\mathbf{62.11 \pm 1.34}$ & \underline{$63.58 \pm 1.37$}
& $\mathbf{66.22 \pm 0.45}$ & $\mathbf{67.91 \pm 0.55}$
& $\mathbf{72.32 \pm 1.64}$ & $\mathbf{62.10 \pm 3.17}$ \\

$H_{\text{ablation}}$
& $59.84 \pm 0.50^\dagger$ & $62.51 \pm 1.61$
& $65.60 \pm 0.63$ & $\underline{67.72 \pm 0.95}$
& $64.24 \pm 7.03$ & $50.29 \pm 10.00$ \\

\cmidrule(lr){1-7}

$H_{\text{sequential}}$
& $61.40 \pm 0.58$ & $61.65 \pm 0.56^\dagger$
& $64.75 \pm 0.60^\dagger$ & $67.14 \pm 0.60$
& $64.44 \pm 2.72^\dagger$ & $44.23 \pm 1.48^\dagger$ \\

$H_{\text{parallel}}$
& \underline{$61.92 \pm 0.89$} & $\mathbf{64.87 \pm 0.70}$
& \underline{$66.14 \pm 0.54$} & $\underline{67.72 \pm 0.64}$
& \underline{$69.03 \pm 1.96^\dagger$} & \underline{$50.79 \pm 1.28^\dagger$} \\

\bottomrule
\end{tabular}
\caption{Comparison of models' $\text{Dynamic}_{\text{max}}$ AULC performance. Best results are \textbf{bolded} and second-best are \underline{underlined}. The $\dagger$ symbol denotes that the model's performance is statistically significantly different to the best model ($p < 0.05$, two-tailed paired t-test).}
\label{tab:in-domain-max-AULC}
\end{table*}

\begin{table*}[t]
\centering
\small
\setlength{\tabcolsep}{4pt}
\begin{tabular}{lcccccccccccc}
\toprule
& \multicolumn{2}{c}{WIKI$\rightarrow$CMV}
& \multicolumn{2}{c}{WIKI$\rightarrow$GITHUB}
& \multicolumn{2}{c}{CMV$\rightarrow$WIKI}
& \multicolumn{2}{c}{CMV$\rightarrow$GITHUB}
& \multicolumn{2}{c}{GITHUB$\rightarrow$WIKI}
& \multicolumn{2}{c}{GITHUB$\rightarrow$CMV} \\
\cmidrule(lr){2-3} \cmidrule(lr){4-5} \cmidrule(lr){6-7}
\cmidrule(lr){8-9} \cmidrule(lr){10-11} \cmidrule(lr){12-13}

Model
& F1 & AUPR
& F1 & AUPR
& F1 & AUPR
& F1 & AUPR
& F1 & AUPR
& F1 & AUPR \\
\midrule
Random
& $50.00$ & $50.00$
& $50.00$ & $23.33$
& $50.00$ & $50.00$
& $50.00$ & $23.33$
& $50.00$ & $50.00$
& $50.00$ & $50.00$\\
\hline
CRAFT\textsuperscript{\ref{fn:CRAFT}} 
& $49.33^\dagger$ & $60.33^\dagger$
& $61.03$ & $41.24^\dagger$
& $52.90^\dagger$ & $58.80^\dagger$
& $50.93^\dagger$ & $32.75^\dagger$
& -- & --
& -- & -- \\

PLM
& \underline{$54.12$} & $\mathbf{64.13}$
& $59.40^\dagger$ & $43.42$
& \underline{$60.34$} & \underline{$64.27$}
& $57.10^\dagger$ & $43.61$
& $50.22$ & \underline{$59.64$}
& \underline{$36.70^\dagger$} & $\mathbf{58.57}$ \\

$H_{\text{ablation}}$
& $51.94^\dagger$ & $62.97$
& $61.27$ & $42.15^\dagger$
& $59.69$ & $63.54^\dagger$
& $56.55^\dagger$ & $40.86$
& $47.13$ & $57.91$
& $36.37^\dagger$ & $57.23$ \\

\hline

$H_{\text{sequential}}$
& $52.31^\dagger$ & $62.11$
& $\mathbf{63.27}$ & \underline{$48.53$}
& $59.32^\dagger$ & $63.43^\dagger$
& $\mathbf{61.02}$ & $\mathbf{44.35}$
& $\mathbf{57.05}$ & $57.14$
& $\mathbf{42.60}$ & \underline{$57.65$} \\

$H_{\text{parallel}}$
& $\mathbf{56.16}$ & \underline{$63.83$}
& \underline{$62.42$} & $\mathbf{49.45}$
& $\mathbf{61.39}$ & $\mathbf{64.53}$
& \underline{$57.33^\dagger$} & \underline{$43.85$}
& \underline{$51.37$} & $\mathbf{63.50}$
& $35.18^\dagger$ & $57.53^\dagger$ \\

\bottomrule
\end{tabular}
\caption{Cross-dataset evaluation of $\text{Dynamic}_{\text{max}}$ AULC using Macro-F1 (F1) and AUPRC (AUPR), standard deviations are omitted for brevity. Each configuration $D1\rightarrow D2$ denotes training on $D1$ and testing on $D2$. Best results are \textbf{bolded} and second-best are \underline{underlined}. The $\dagger$ symbol denotes that the model's performance is statistically significantly different to the best model ($p < 0.05$, two-tailed paired t-test).}
\label{tab:cross-domain-max-AULC}
\end{table*}

\subsection{Model Performance Across Data Regime}\label{sec:appendix-learning-curve}
\begin{figure}[h]
    \centering
    \includegraphics[width=0.9\linewidth]{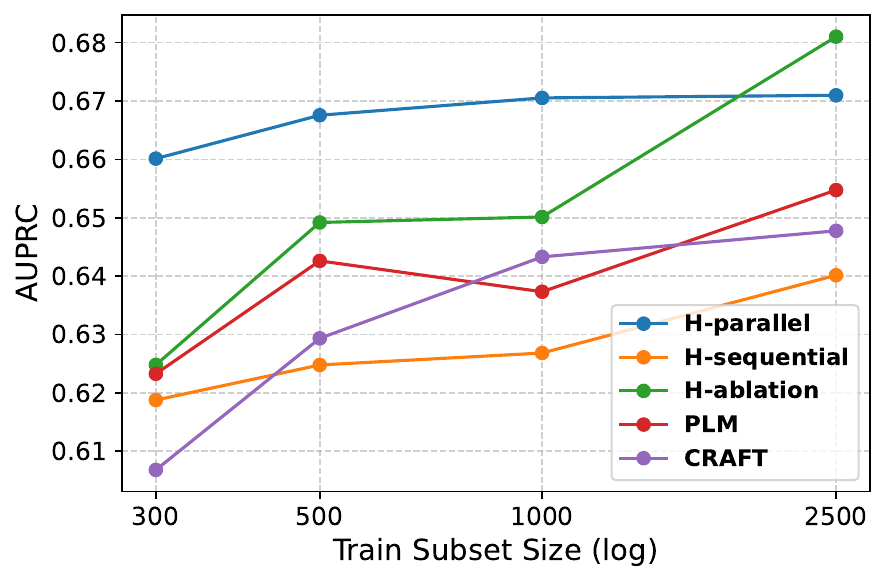}
    \caption{Comparison of models' $\text{Dynamic}_{\text{mean}}$ AUPRC learning curves on WIKI, plotted against training subset size.}
    \label{fig:WIKI-WIKI-mean-AUPRC}
\end{figure}

\begin{figure}[h]
    \centering
    \includegraphics[width=0.9\linewidth]{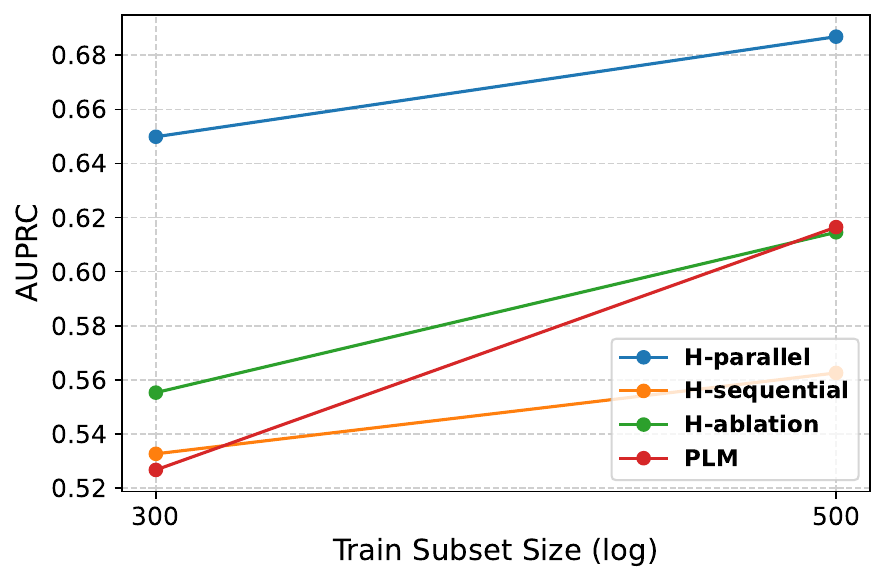}
    \caption{Comparison of models' $\text{Dynamic}_{\text{mean}}$ AUPRC learning curves on GITHUB, plotted against training subset size.}
    \label{fig:GITHUB-GITHUB-mean-AUPRC}
\end{figure}

Figures~\ref{fig:WIKI-WIKI-mean-AUPRC} and~\ref{fig:GITHUB-GITHUB-mean-AUPRC} present the $\text{Dynamic}_{\text{mean}}$ AUPRC learning curves on WIKI and GITHUB, respectively. They follow a similar trend to CMV (Figure~\ref{fig:CMV-CMV-mean-AUPRC}), where $H_{\text{parallel}}$ demonstrates strong performance in low-data regimes, although it can be outperformed by text-based models when larger amounts of data are available. Specifically, $H_{\text{parallel}}$ exhibits a smaller decline in performance as the number of training samples decreases. By leveraging pragmatic signals, it achieves higher performance with just 300 samples than other models do with 500 or 1000, significantly reducing the amount of data required to reach comparable results.

\begin{figure*}[h]
    \centering
    \includegraphics[width=0.9\linewidth]{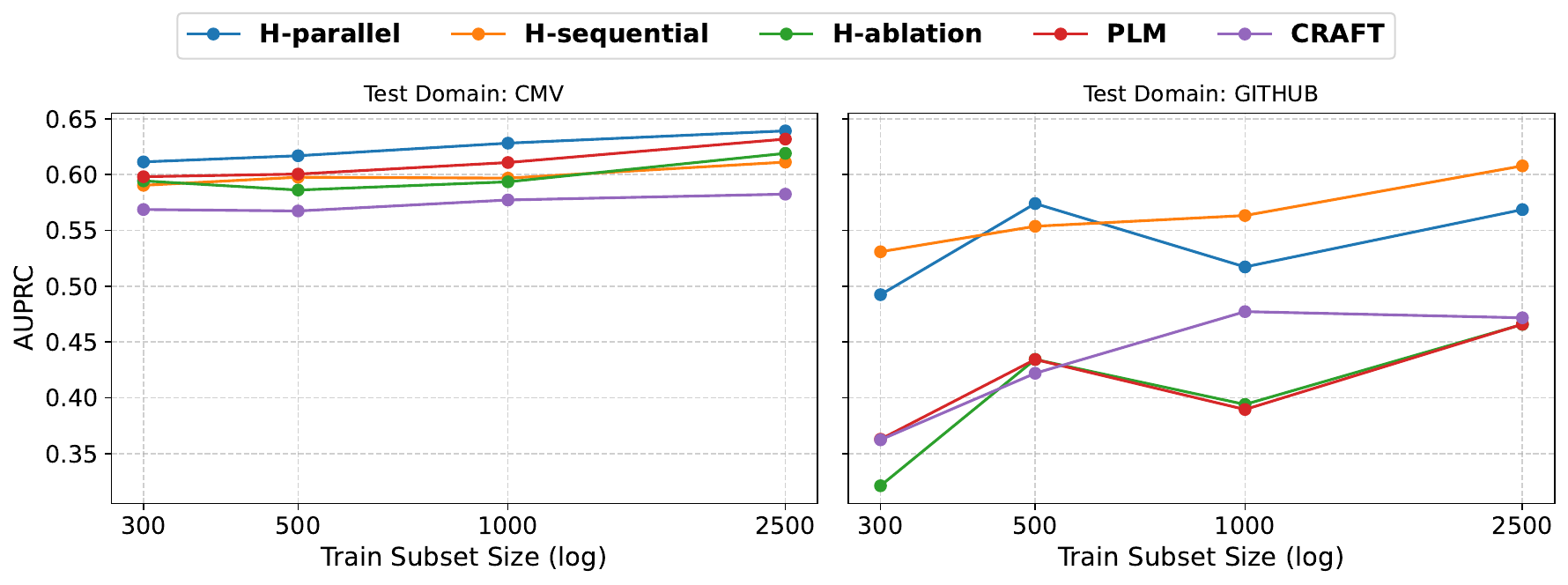}
    \caption{Comparison of models' cross-dataset $\text{Dynamic}_{\text{mean}}$ AUPRC learning curves when trained on WIKI, plotted against training subset size.}
    \label{fig:WIKI-cross-mean-AUPRC}
\end{figure*}

\begin{figure*}[h]
    \centering
    \includegraphics[width=0.9\linewidth]{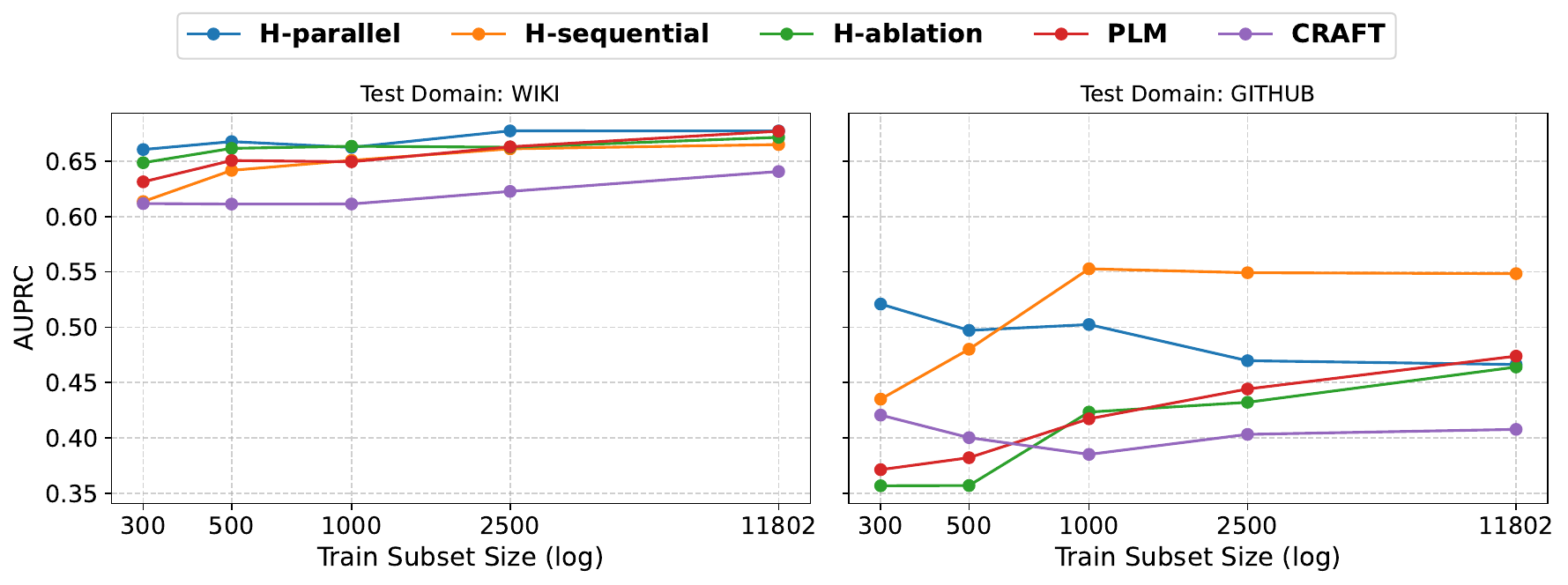}
    \caption{Comparison of models' cross-dataset $\text{Dynamic}_{\text{mean}}$ AUPRC learning curves when trained on CMV, plotted against training subset size.}
    \label{fig:CMV-cross-mean-AUPRC}
\end{figure*}

\begin{figure*}[h]
    \centering
    \includegraphics[width=0.9\linewidth]{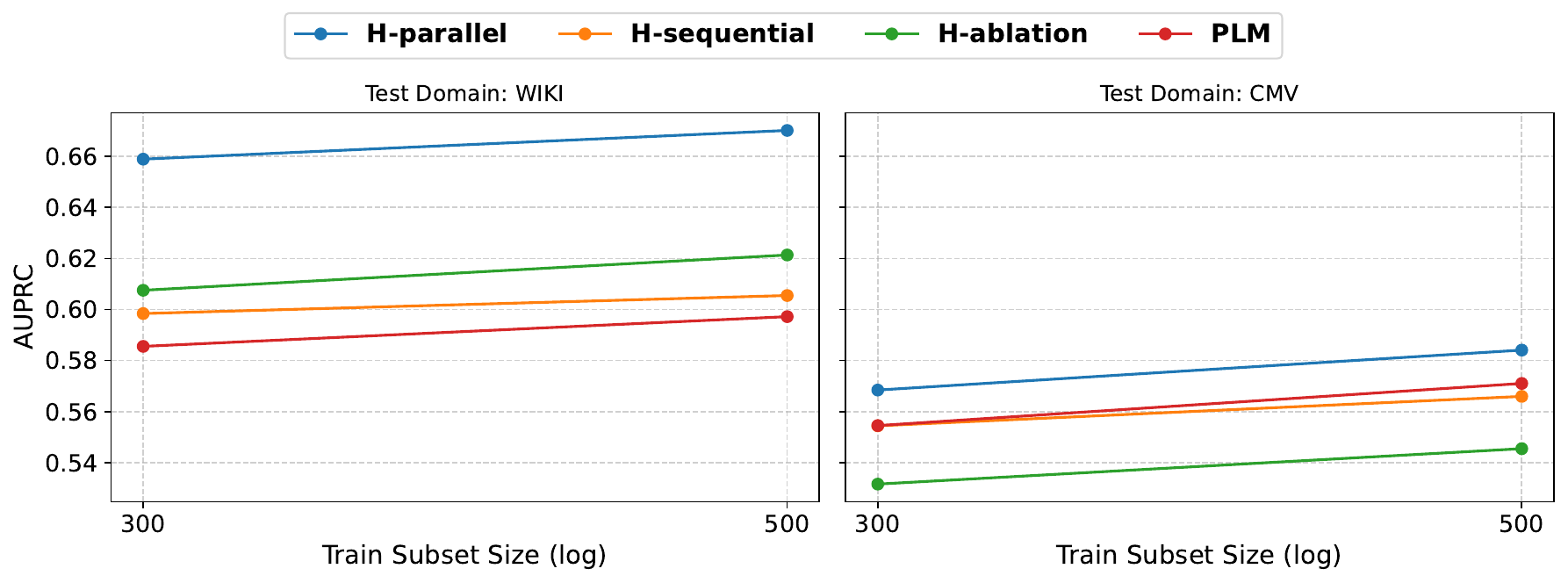}
    \caption{Comparison of models' cross-dataset $\text{Dynamic}_{\text{mean}}$ AUPRC learning curves when trained on GITHUB, plotted against training subset size.}
    \label{fig:GITHUB-cross-mean-AUPRC}
\end{figure*}

In addition, \Cref{fig:WIKI-cross-mean-AUPRC,fig:CMV-cross-mean-AUPRC,fig:GITHUB-cross-mean-AUPRC} present the learning curves in cross-dataset settings for models trained on WIKI, CMV, and GITHUB, respectively. As with the in-domain results, SA signals generally provide greater benefits in low-data regimes. The cross-domain performance of $H_{\text{parallel}}$ on GITHUB tends to fluctuate as the number of training samples increases. In particular, when trained on CMV, its performance eventually degrades to the level of the text-based models (PLM and $H_{\text{ablation}}$) as the training size reaches 11,802. This suggests that $H_{\text{parallel}}$ may require further tuning with higher regularization to more effectively utilize SA features. In contrast, $H_{\text{sequential}}$, which prioritizes SA features, is less prone to overfitting the source domain and demonstrates robust generalizability to GITHUB.

\footnotetext[4]{\label{fn:epoch}10 epochs on CMV and 15 epochs on Wiki and GITHUB.}

\subsection{Per-Timestep Model Performance}\label{sec:appendix-timestep-performance}
\begin{figure*}[t]
    \centering
    \begin{subfigure}[b]{0.495\textwidth}
        \centering
        \includegraphics[width=\textwidth]{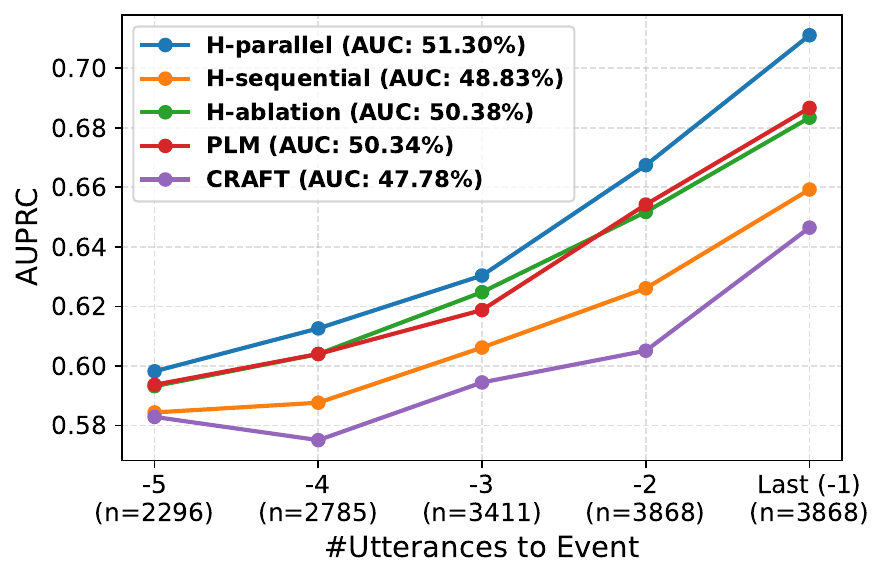}
        \caption{300 Train Subset}
    \end{subfigure}
    \hfill
    \begin{subfigure}[b]{0.495\textwidth}
        \centering
        \includegraphics[width=\textwidth]{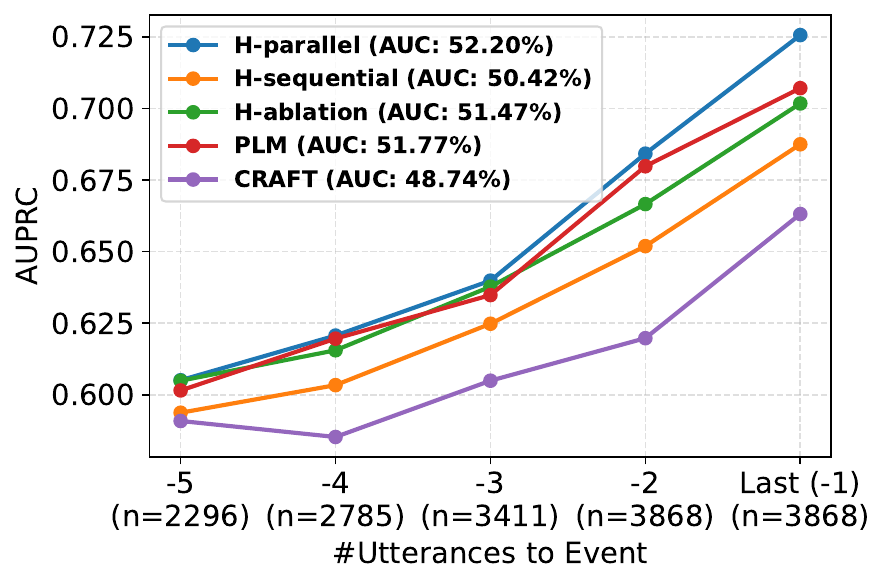}
        \caption{500 Train Subset}
    \end{subfigure}
    \hfill
    \begin{subfigure}[b]{0.495\textwidth}
        \centering
        \includegraphics[width=\textwidth]{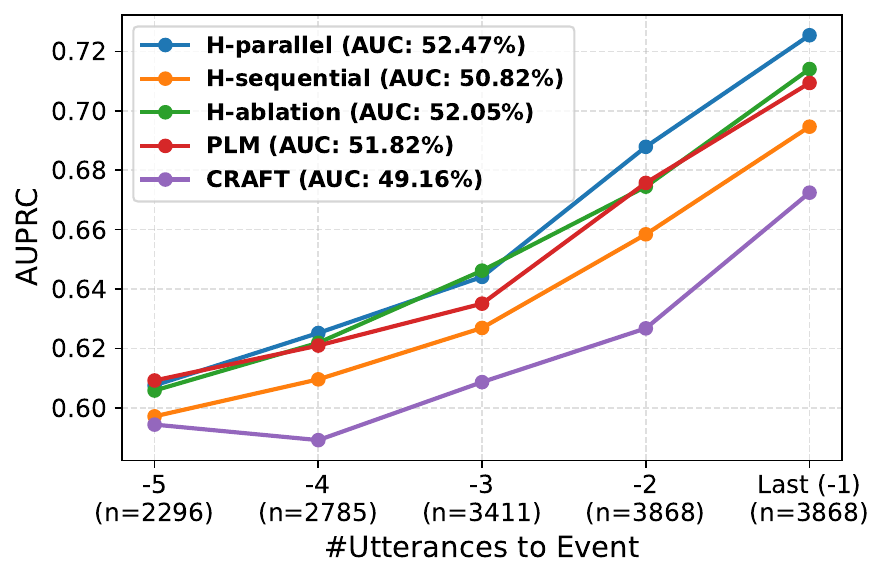}
        \caption{1000 Train Subset}
    \end{subfigure}
    \hfill
    \begin{subfigure}[b]{0.495\textwidth}
        \centering
        \includegraphics[width=\textwidth]{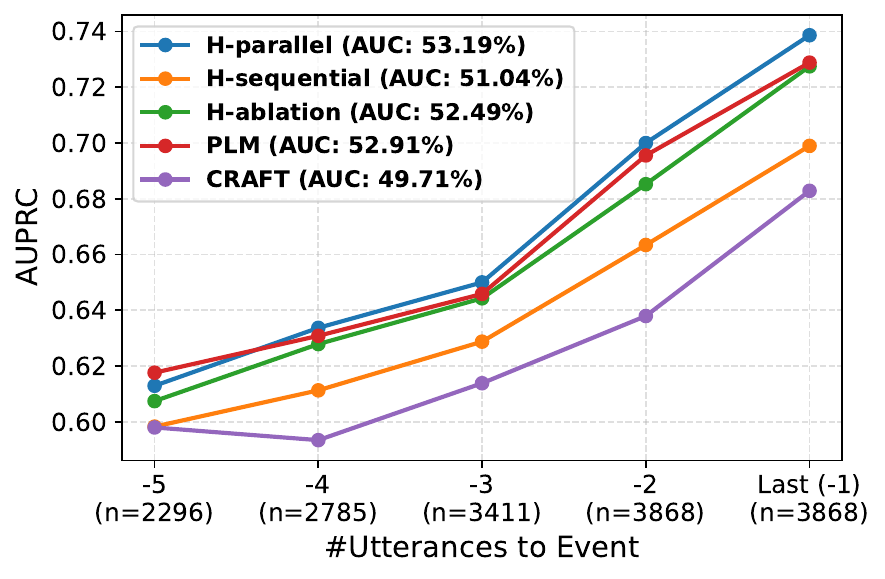}
        \caption{2500 Train Subset}
    \end{subfigure}
    \hfill
    \begin{subfigure}[b]{0.495\textwidth}
        \centering
        \includegraphics[width=\textwidth]{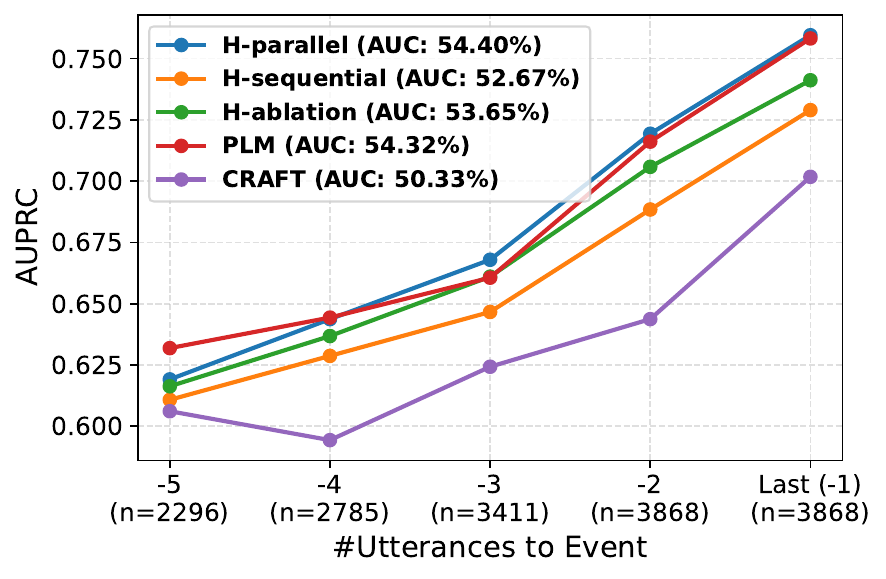}
        \caption{11802 Train Subset}
    \end{subfigure}

    \caption{Model AUPRC performance on CMV is plotted as dialogues progress toward derailment or conversation end, under varying training data sizes and averaged across five random seeds. The number of test-set dialogues contributing to each time step (n) is shown on the x-axis ticks, as some dialogues contain fewer than five utterances. Areas under the curves are reported in the legend. $H_{\text{parallel}}$ achieves the best performance across time steps, particularly in the low-data regime.}
    \label{fig:timestep-performance-CMV-p2}
\end{figure*}

\begin{figure*}[t]
    \centering

    \begin{subfigure}[b]{0.495\textwidth}
        \centering
        \includegraphics[width=\textwidth]{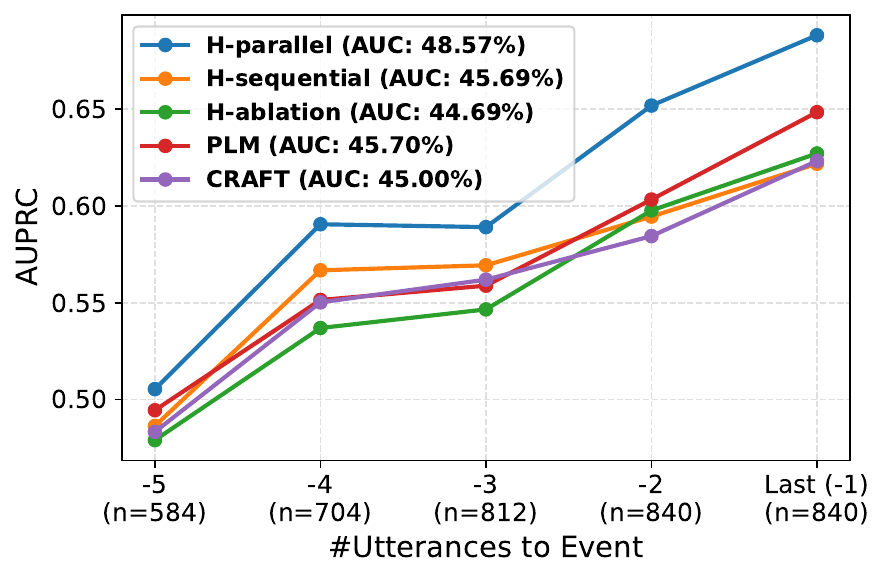}
        \caption{300 Train Subset}
    \end{subfigure}
    \hfill
    \begin{subfigure}[b]{0.495\textwidth}
        \centering
        \includegraphics[width=\textwidth]{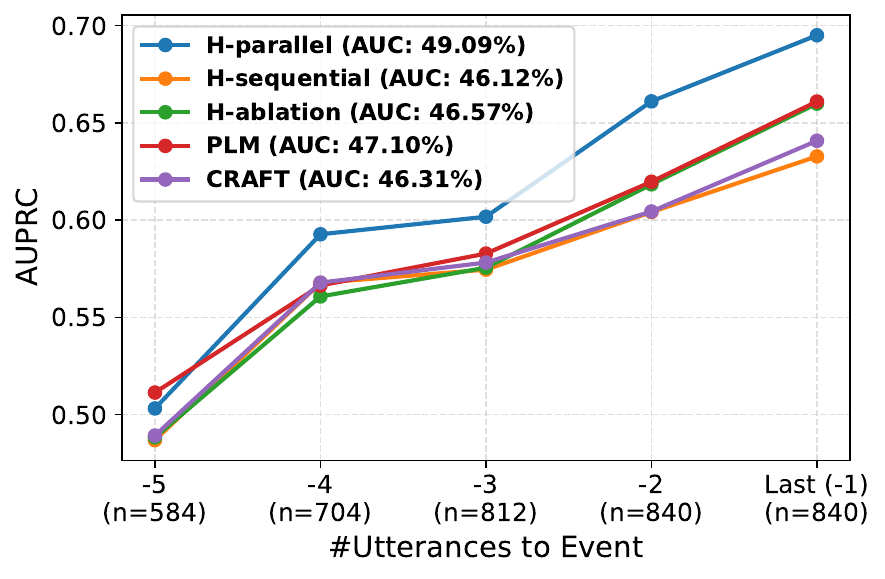}
        \caption{500 Train Subset}
    \end{subfigure}
    \hfill
    \begin{subfigure}[b]{0.495\textwidth}
        \centering
        \includegraphics[width=\textwidth]{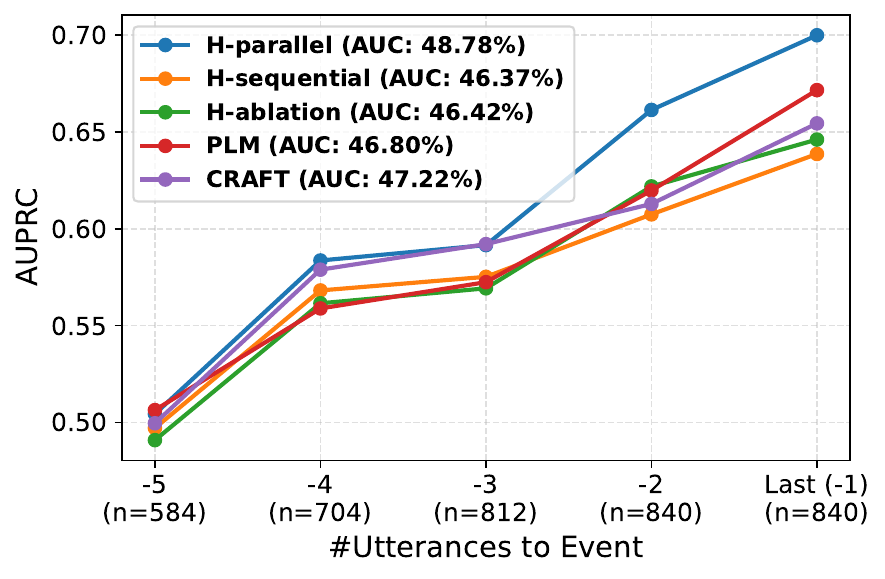}
        \caption{1000 Train Subset}
    \end{subfigure}
    \hfill
    \begin{subfigure}[b]{0.495\textwidth}
        \centering
        \includegraphics[width=\textwidth]{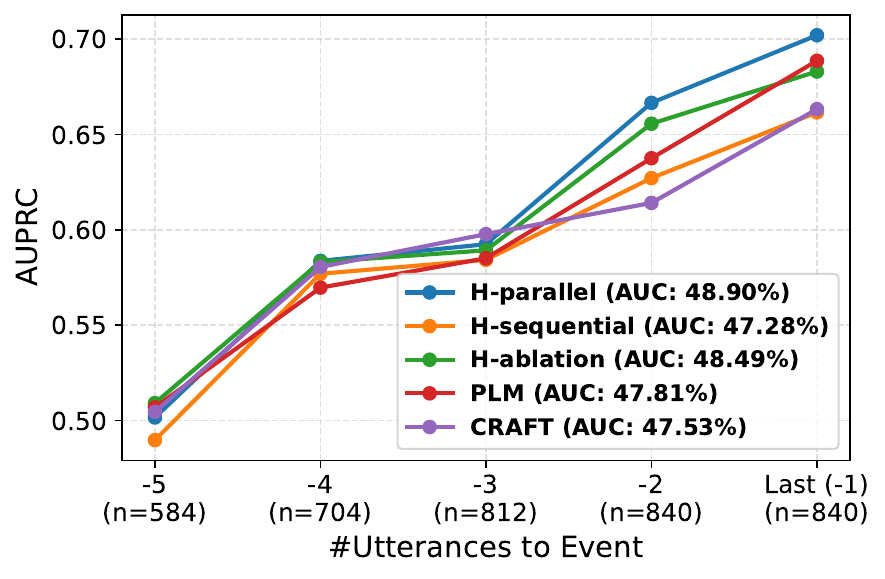}
        \caption{2500 Train Subset}
    \end{subfigure}

    \caption{Model AUPRC performance on WIKI is plotted as dialogues progress toward derailment or conversation end, under varying training data sizes and averaged across five random seeds. The number of test-set dialogues contributing to each time step (n) is shown on the x-axis ticks, as some dialogues contain fewer than five utterances. Areas under the curves are reported in the legend. $H_{\text{parallel}}$ achieves the best performance across time steps, particularly in the low-data regime.}
    \label{fig:timestep-performance-WIKI}
\end{figure*}

\begin{figure*}[t]
    \centering

    \begin{subfigure}[b]{0.495\textwidth}
        \centering
        \includegraphics[width=\textwidth]{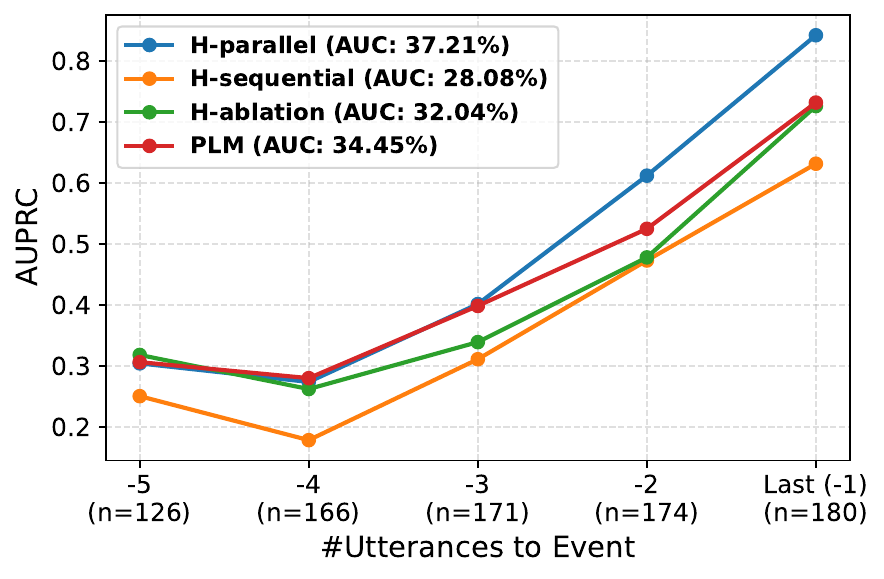}
        \caption{300 Train Subset}
    \end{subfigure}
    \hfill
    \begin{subfigure}[b]{0.495\textwidth}
        \centering
        \includegraphics[width=\textwidth]{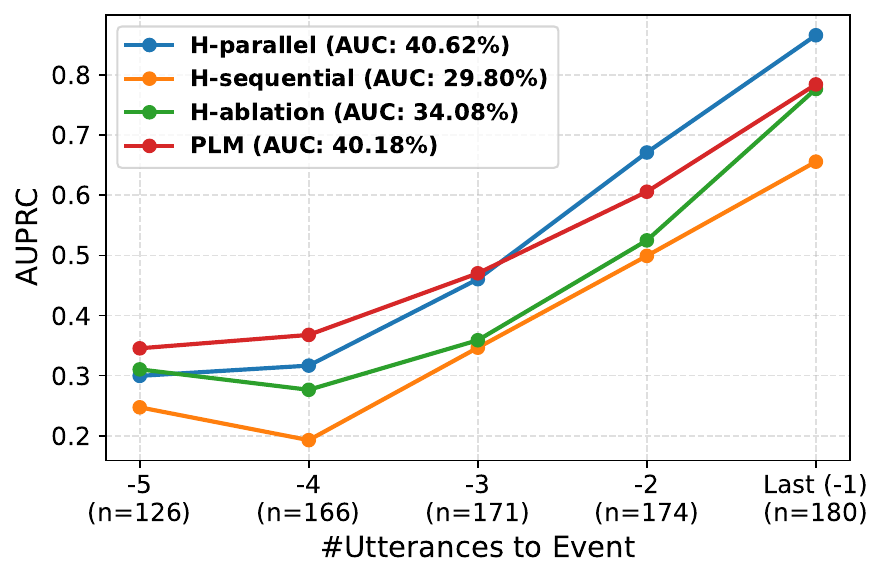}
        \caption{500 Train Subset}
    \end{subfigure}

    \caption{Model AUPRC performance on GITHUB is plotted as dialogues progress toward derailment or conversation end, under varying training data sizes and averaged across five random seeds. The number of test-set dialogues contributing to each time step (n) is shown on the x-axis ticks, as some dialogues contain fewer than five utterances. Areas under the curves are reported in the legend. $H_{\text{parallel}}$ achieves the best performance across time steps, particularly in the low-data regime.}
    \label{fig:timestep-performance-GITHUB}
\end{figure*}

\Cref{fig:timestep-performance-CMV-p2,fig:timestep-performance-WIKI,fig:timestep-performance-GITHUB} present the per-timestep performance of models as dialogues progress toward derailment or the end of the conversation, on CMV, WIKI, and GITHUB, respectively. All datasets exhibit similar trends discussed in Section~\ref{sec:in-domain-performance}. Overall, $H_{\text{parallel}}$ shows the best performance across time steps, particularly in low-data regimes, which generally aligns with the $\text{Dynamic}_{\text{mean}}$ rankings.

\subsection{Robustness to PLM}\label{sec:appendix-bert}
To assess the robustness of our proposed approach across different underlying PLMs, we perform a supplementary experiment using \texttt{bert-base-uncased}~\cite{BERT-Devlin2018} instead of \texttt{RoBERTa-large}, as \texttt{BERT} is also a widely adopted baseline in prior derailment forecasting work~\cite{dynamic-Kementchedjhieva2021}. Specifically, the PLM components in all relevant models (PLM, $H_{\text{ablation}}$, $H_{\text{sequential}}$, and $H_{\text{parallel}}$) are replaced with \texttt{bert-base-uncased}. All hyperparameters remain consistent with prior experiments. This evaluation is conducted using the CMV dataset for training, with the resulting models tested across all datasets. Table~\ref{tab:mean-AULC-bert} 
reports the $\text{Dynamic}_{\text{mean}}$ 
AULC performance.
Because it uses a smaller architecture, BERT generally performs worse than RoBERTa. Regardless, the SA-based models achieve the strongest in-domain and cross-domain performance, showing a consistent trend to the results obtained with RoBERTa-large. This illustrates the robustness of our approach and its ability to improve generalizability upon different underlying PLM.

\begin{table*}[t]
\centering
\small
\setlength{\tabcolsep}{6pt}
\begin{tabular}{lcccccc}
\toprule
& \multicolumn{2}{c}{CMV $\rightarrow$ WIKI} & \multicolumn{2}{c}{CMV} & \multicolumn{2}{c}{CMV $\rightarrow$ GITHUB} \\
\cmidrule(lr){2-3} \cmidrule(lr){4-5} \cmidrule(lr){6-7}
Model 
& MacroF1 & AUPRC
& MacroF1 & AUPRC
& MacroF1 & AUPRC \\
\midrule
Random
& $50.00$ & $50.00$
& $50.00$ & $50.00$
& $50.00$ & $23.33$\\
\hline
CRAFT 
& $53.31 \pm 3.23^\dagger$ & $62.14 \pm 1.46^\dagger$
& $61.97 \pm 0.47^\dagger$ & $62.76 \pm 0.41^\dagger$
& $51.75 \pm 1.47^\dagger$ & $40.09 \pm 2.80$ \\

PLM (BERT)
& $56.90 \pm 0.87^\dagger$ & $60.89 \pm 0.76^\dagger$
& \underline{$62.61 \pm 0.46^\dagger$} & \underline{$64.90 \pm 0.63^\dagger$}
& $45.60 \pm 5.12^\dagger$ & $37.60 \pm 2.09^\dagger$ \\

$H_{\text{ablation}}$
& $55.06 \pm 2.39^\dagger$ & $60.39 \pm 0.60^\dagger$
& $62.18 \pm 0.73^\dagger$ & $63.50 \pm 0.62^\dagger$
& $50.51 \pm 4.22^\dagger$ & $37.59 \pm 2.46^\dagger$ \\

\cmidrule(lr){1-7}

$H_{\text{sequential}}$
& \underline{$58.17 \pm 0.83$} & $\mathbf{64.20 \pm 0.53}$
& $62.58 \pm 0.46^\dagger$ & $64.57 \pm 0.30^\dagger$
& $\mathbf{59.93 \pm 2.21}$ & \underline{$41.62 \pm 2.69$} \\

$H_{\text{parallel}}$
& $\mathbf{58.53 \pm 0.57}$ & \underline{$63.92 \pm 0.25$}
& $\mathbf{63.48 \pm 0.68}$ & $\mathbf{65.61 \pm 0.48}$
& \underline{$54.00 \pm 4.67^\dagger$} & $\mathbf{45.12 \pm 3.95}$ \\
\bottomrule
\end{tabular}
\caption{Robustness to PLM: Comparison of models' $\text{Dynamic}_{\text{mean}}$ AULC performance when bert-base-uncased is used as the base model. Models are trained on CMV and tested on all datasets. Best results are \textbf{bolded} and second-best are \underline{underlined}. The $\dagger$ symbol denotes that the model's performance is statistically significantly different to the best model ($p < 0.05$, two-tailed paired t-test). $H_{\text{parallel}}$ achieves strongest in-domain performance (on CMV), while SA-based models ($H_{\text{parallel}}$ and $H_{\text{sequential}}$) achieve strongest cross-dataset performance (CMV $\rightarrow$ WIKI and CMV $\rightarrow$ GITHUB).}
\label{tab:mean-AULC-bert}
\end{table*}

\subsection{Robustness to Model Architecture: GraphNLI}\label{sec:appendix-graphNLI}
To assess the robustness of the proposed approach to the model architecture, we conduct a supplementary experiment running a graph-based context-aware model GraphNLI~\cite{GraphNLI-Agarwal2023}. We follow its original implementation and the hyperparameters for the hate speech detection task. Minor adjustments to the hyperparameters are made to adapt the model to the derailment forecasting task: increasing the number of training epochs for smaller training sets (same as PLM/H-ablation training epochs as reported in Table~\ref{tab:epoch}), and performing a root-seeking graph walk (setting $p=1$ and $\gamma=0.2$) as our task focuses on a single conversation rather than a branching conversation.

To focus on the effect of SA integration, we trained a variant GraphNLI-SA, which includes a SA classification head on top of its PLM encoder, using the same auxiliary SA classification objective and two-phased training procedure as H-parallel. While GraphNLI uses a weaker PLM (\texttt{distilroberta-base}~\cite{DistilBERT-Sanh2019}), the full PLM is trained during the intermediate training phase in order to effectively learn the complex SA classification task. The number of finetune epochs is consistent with $H_{parallel}$ reported in Table~\ref{tab:epoch}, where the intermediate phase used an increased number of epochs to accomodate the weaker PLM (25/15/10/5/2 epochs for 300/500/1000/2500/11802 training samples respectively). Other hyperparameters remain consistent with GraphNLI.

Table~\ref{tab:graphNLI} reports the resulting model performance. Despite the architectural differences between GraphNLI and our hierarchical models, the results are consistent with previous findings: incorporating SA supervision consistently improves in-domain and cross-domain performance. 

While different model architectures are possible, they largely build on top of PLMs as utterance encoders, with their primary differences lying in how conversational context is modeled and integrated. The consistent gains observed across both GraphNLI and our hierarchical framework suggest the performance gains arise from the additional SA supervision rather than the specific hierarchical architecture, demonstrating that the proposed approach is compatible with alternative backbone models.

\begin{table*}[t]
\centering
\small
\setlength{\tabcolsep}{4pt}
\begin{tabular}{llcccccc}
\toprule
&
& \multicolumn{2}{c}{Test: WIKI}
& \multicolumn{2}{c}{Test: CMV}
& \multicolumn{2}{c}{Test: GITHUB} \\
\cmidrule(lr){3-4} \cmidrule(lr){5-6} \cmidrule(lr){7-8}
Train & Model
& MacroF1 & AUPRC
& MacroF1 & AUPRC
& MacroF1 & AUPRC \\
\midrule

\multirow{2}{*}{WIKI}
& GraphNLI
& $61.40$
& $65.86$
& $41.24^\dagger$
& $58.43^\dagger$
& $63.53$
& $52.22$ \\

& GraphNLI-SA
& $\mathbf{63.03}$
& $\mathbf{67.45}$
& $\mathbf{46.86}$
& $\mathbf{62.14}$
& $\mathbf{66.00}$
& $\mathbf{54.87}$ \\

\midrule

\multirow{2}{*}{CMV}
& GraphNLI
& $56.31^\dagger$
& $63.77$
& $63.28^\dagger$
& $65.47^\dagger$
& $57.06$
& $44.94$ \\

& GraphNLI-SA
& $\mathbf{58.97}$
& $\mathbf{64.64}$
& $\mathbf{64.05}$
& $\mathbf{66.50}$
& $\mathbf{58.55}$
& $\mathbf{46.79}$ \\

\midrule

\multirow{2}{*}{GITHUB}
& GraphNLI
& $45.55$
& $59.16^\dagger$
& $33.68$
& $54.52$
& $70.30$
& $60.54$ \\

& GraphNLI-SA
& $\mathbf{48.95}$
& $\mathbf{62.40}$
& $\mathbf{34.01}$
& $\mathbf{55.95}$
& $\mathbf{73.52}$
& $\mathbf{65.12}$ \\

\bottomrule
\end{tabular}
\caption{Ablation study on $\text{Dynamic}_{\text{mean}}$ performance of GraphNLI. The $\dagger$ symbol denotes that the model's performance is statistically significantly different to the best model ($p < 0.05$, two-tailed paired t-test). Standard deviations are omitted for brevity. The best performance is \textbf{bolded}. SA integration consistently improves both in-domain and cross-domain performance.}
\label{tab:graphNLI}
\end{table*}

\subsection{Robustness to SA Extraction}\label{sec:appendix-SA-robustness}
\begin{figure*}[htbp]
\centering
\begin{forest}
  for tree={
    node options={mynode},
    edge={thick, draw=black},
    parent anchor=south,
    child anchor=north,
    s sep=2mm, 
    l=1.5cm   
  }
  [Expressive
    [Positive
      [{Forgive, Praise,\\ Congratulate}]
      [Thank]
    ]
    [Negative
      [Apologize]
      [{Complain, Criticize,\\ Insult, Deplore}]
    ]
    [{Greet, Farewell}]
    [Wish]
  ]
\end{forest}
\caption{Expressive Speech Act Taxonomy}
\label{fig:speech-act-expressive}
\end{figure*}
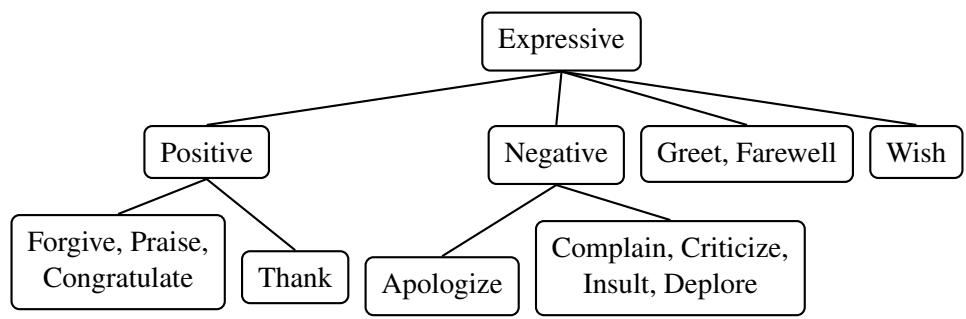

To assess the robustness of the proposed approach to the underlying SA extraction process, we conduct a supplementary experiment using a more coarse-grained speech act (SA) taxonomy. While the main experiments employ a fine-grained set of 51 speech acts (including ``other''), these acts are organized according to the prior taxonomy proposed by \citet{SAOpenDomain-Compagno2018}, enabling hierarchical merging where necessary.
Specifically, from \Cref{fig:speech-act-assertive,fig:speech-act-directive,fig:speech-act-commissive,fig:speech-act-expressive}, speech acts belonging to the same node in the taxonomy are collapsed into a single class, resulting in 17 speech act classes and an additional ``other'' label.

As the proposed approach is primarily intended to improve performance in low-data regimes, and because speech act distributions and interaction patterns may vary across datasets, we conduct this supplementary experiment across the 300-training-sample setting, as oppose to across a dataset. All hyperparameters remain consistent with prior experiments. We compare the performance of our main model, $H_{\text{parallel}}$, using the merged SA taxonomy, against the baselines, and against the original fine-grained SA taxonomy setting. 

Table~\ref{tab:300-saRobustness} reports the resulting model performance. We observe that $H_{\text{parallel}}$ is robust to the specific SA extraction used. Their in-domain performance (dialogonals in the table) remains highly comparable, with the largest absolute discrepancy being a minor 0.56\% AUPRC and 1.59\% Macro-F1 on GITHUB. Cross-dataset performance displays similar consistency, particularly in terms of AUPRC, where the maximum deviation is 2.83\% (CMV $\rightarrow$ GITHUB), with the lower outperforming best baseline by 7.2\%. Conversely, MacroF1 exhibits higher instability and greater divergence, peaking at a 6.91\% difference for GITHUB $\rightarrow$ CMV. This variance could be driven by a reliance on source-tuned classification thresholds, as their underlying AUPRC scores differ by only 0.32\%.

\begin{table*}[t]
\centering
\small
\setlength{\tabcolsep}{4pt}
\begin{tabular}{llcccccc}
\toprule
& 
& \multicolumn{2}{c}{Test: WIKI} 
& \multicolumn{2}{c}{Test: CMV} 
& \multicolumn{2}{c}{Test: GITHUB} \\
\cmidrule(lr){3-4} \cmidrule(lr){5-6} \cmidrule(lr){7-8}
Train & Model
& MacroF1 & AUPRC 
& MacroF1 & AUPRC 
& MacroF1 & AUPRC \\
\midrule

& Random
& $50.00$ & $50.00$
& $50.00$ & $50.00$
& $50.00$ & $23.33$\\
\hline
\multirow{5}{*}{WIKI}
& CRAFT
& $58.57^\dagger$ & $60.68^\dagger$
& $50.61$ & $56.87^\dagger$
& $58.93$ & $36.23^\dagger$ \\

& PLM
& $61.64$ & $62.33^\dagger$
& $\mathbf{51.85}$ & $59.81^\dagger$
& $55.06$ & $36.29^\dagger$ \\

& $H_{\text{ablation}}$
& $58.59^\dagger$ & $62.48^\dagger$
& $49.95$ & $59.43^\dagger$
& $54.21^\dagger$ & $32.12^\dagger$ \\

\cmidrule(lr){2-8}
& $H_{parallel,18}$
& \underline{$62.39$} & \underline{$65.85$}
& $49.48$ & $\mathbf{61.58}$
& $\mathbf{64.18}$ & $\mathbf{50.74}$ \\

& $H_{parallel,51}$
& $\mathbf{62.51}$ & $\mathbf{66.01}$
& \underline{$50.89$} & \underline{$61.13$}
& \underline{$62.14$} & \underline{$49.24$} \\

& $|\Delta|$
& 0.12 & 0.16
& 1.41 & 0.45
& 2.04 & 1.50 \\

\midrule

\multirow{5}{*}{CMV}
& CRAFT
& $50.42$ & $61.18^\dagger$
& $59.47^\dagger$ & $60.06^\dagger$
& $47.60^\dagger$ & $42.06$ \\

& PLM
& $58.50$ & $63.14$
& $62.15$ & $62.48^\dagger$
& $53.97$ & $37.12^\dagger$ \\

& $H_{\text{ablation}}$
& $56.86$ & $64.87^\dagger$
& $62.51$ & $63.76$
& $53.23$ & $35.67$ \\

\cmidrule(lr){2-8}

& $H_{parallel,18}$
& $\mathbf{59.28}$ & \underline{$64.94$}
& \underline{$63.00$} & \underline{$64.95$}
& \underline{$54.99$} & \underline{$49.26$} \\

& $H_{parallel,51}$
& \underline{$59.22$} & $\mathbf{66.06}$
& $\mathbf{63.03}$ & $\mathbf{65.22}$
& $\mathbf{60.36}$ & $\mathbf{52.09}$ \\

& $|\Delta|$
& 0.06 & 1.12
& 0.03 & 0.27
& 5.40 & 2.83 \\

\midrule

\multirow{4}{*}{GITHUB}
& PLM
& $44.62^\dagger$ & $58.56^\dagger$
& $34.70$ & $55.45$
& $69.24$ & $52.67^\dagger$ \\

& $H_{\text{ablation}}$
& \underline{$53.27$} & $60.75^\dagger$
& $\mathbf{38.23}$ & $53.16^\dagger$
& $63.92$ & $55.52$ \\

\cmidrule(lr){2-8}

& $H_{parallel,18}$
& $\mathbf{55.35}$ & $\mathbf{66.21}$
& \underline{$34.87$} & \underline{$56.09$}
& $\mathbf{70.87}$ & \underline{$64.43$} \\

& $H_{parallel,51}$
& $48.44^\dagger$ & \underline{$65.89$}
& $33.92$ & $\mathbf{56.84}$
& \underline{$69.28$} & $\mathbf{64.99}$ \\

& $|\Delta|$
& 6.91 & 0.32
& 0.95 & 0.75
& 1.59 & 0.56 \\

\bottomrule
\end{tabular}
\caption{Robustness to SA Extraction: Comparison of $\text{Dynamic}_{\text{mean}}$ performance between different models under 300 training samples, where $H_{\text{parallel}}$ uses a fine-grained ($H_{parallel,51}$) and a coarse-grained ($H_{parallel,18}$) taxonomy. $|\Delta|$ denotes the absolute difference between their performance. The $\dagger$ symbol denotes that the model's performance is statistically significantly different to the best model ($p < 0.05$, two-tailed paired t-test). Standard deviations are omitted for brevity. The best performance is \textbf{bolded}, and second-best \underline{underlined}. Model performance is relatively robust to SA extraction used, with the two taxonomies achieving comparable performance to each other.}
\label{tab:300-saRobustness}
\end{table*}

\subsection{Validation Set Result}\label{sec:appendix-validation}
For reproducibility purpose, Tables~\ref{tab:mean-AULC-validation} and~\ref{tab:max-AULC-validation} report the validation set AULC performance of the models.

\begin{table*}[t]
\centering
\small
\setlength{\tabcolsep}{6pt}
\begin{tabular}{lcccccc}
\toprule
& \multicolumn{2}{c}{WIKI} & \multicolumn{2}{c}{CMV} & \multicolumn{2}{c}{GITHUB} \\
\cmidrule(lr){2-3} \cmidrule(lr){4-5} \cmidrule(lr){6-7}
Model 
& MacroF1 & AUPRC
& MacroF1 & AUPRC
& MacroF1 & AUPRC \\
\midrule
CRAFT
& $60.55 \pm 0.70$ & $58.42 \pm 0.49$
& $61.87 \pm 0.30$ & $63.57 \pm 0.09$
& -- & -- \\

PLM
& $62.41 \pm 0.75$ & $61.97 \pm 0.84$
& $65.40 \pm 0.21$ & $66.01 \pm 0.26$
& $70.43 \pm 2.83$ & $43.79 \pm 3.17$ \\

$H_{\text{ablation}}$
& $60.83 \pm 1.19$ & $61.03 \pm 1.37$
& $64.72 \pm 0.42$ & $66.17 \pm 0.30$
& $69.51 \pm 2.29$ & $50.75 \pm 4.87$ \\

$H_{\text{sequential}}$
& $60.84 \pm 0.86$ & $60.65 \pm 0.97$
& $62.88 \pm 0.34$ & $65.87 \pm 0.44$
& $77.25 \pm 2.09$ & $58.65 \pm 1.88$ \\

$H_{\text{parallel}}$
& $61.87 \pm 0.37$ & $62.16 \pm 0.58$
& $65.44 \pm 0.22$ & $67.24 \pm 0.28$
& $71.78 \pm 1.75$ & $55.47 \pm 1.45$ \\

\bottomrule
\end{tabular}
\caption{Models' validation set $\text{Dynamic}_{\text{mean}}$ AULC performance for reproducibility purpose. Macro-F1 uses threshold tuned on the validation-set.}
\label{tab:mean-AULC-validation}
\end{table*}

\begin{table*}[t]
\centering
\small
\setlength{\tabcolsep}{6pt}
\begin{tabular}{lcccccc}
\toprule
& \multicolumn{2}{c}{WIKI} & \multicolumn{2}{c}{CMV} & \multicolumn{2}{c}{GITHUB} \\
\cmidrule(lr){2-3} \cmidrule(lr){4-5} \cmidrule(lr){6-7}
Model 
& MacroF1 & AUPRC
& MacroF1 & AUPRC
& MacroF1 & AUPRC \\
\midrule
CRAFT
& $60.71 \pm 0.58$ & $58.81 \pm 0.56$
& $62.46 \pm 0.66$ & $65.14 \pm 0.61$
& -- & -- \\

PLM
& $62.79 \pm 1.39$ & $62.10 \pm 1.13$
& $66.03 \pm 0.24$ & $68.27 \pm 0.65$
& $69.38 \pm 2.62$ & $46.46 \pm 5.02$ \\

$H_{\text{ablation}}$
& $60.14 \pm 1.01$ & $60.84 \pm 1.75$
& $64.80 \pm 0.46$ & $66.96 \pm 0.73$
& $61.74 \pm 3.78$ & $39.88 \pm 7.80$ \\

$H_{\text{sequential}}$
& $60.23 \pm 0.66$ & $60.13 \pm 1.05$
& $63.84 \pm 0.43$ & $67.90 \pm 0.42$
& $67.73 \pm 3.59$ & $42.29 \pm 5.66$ \\

$H_{\text{parallel}}$
& $61.72 \pm 0.83$ & $62.02 \pm 0.75$
& $65.60 \pm 0.29$ & $66.19 \pm 0.54$
& $65.44 \pm 1.48$ & $40.93 \pm 1.49$ \\

\bottomrule
\end{tabular}
\caption{Models' validation set $\text{Dynamic}_{\text{max}}$ AULC performance. Macro-F1 uses threshold tuned on the validation-set.}
\label{tab:max-AULC-validation}
\end{table*}

\begin{figure*}[htbp]
\centering
\begin{forest}
  for tree={
    node options={mynode},
    edge={thick, draw=black},
    parent anchor=south,
    child anchor=north,
    s sep=2mm,
    l=1.5cm   
  }
  [Assertive
    [Independent from previous
      [Strong Strength
        [{Assert, Boast, Swear,\\
         Insist, Emphasise}]
      ]
      [Medium Strength
        [{State, Inform, Report,\\
         Conclude, Quote,\\
         Claim, Remind}]
      ]
      [Weak Strength
        [{Predict,\\ Hypothesise,\\ Suggest}]
      ]
    ]
    [Dependent from previous
      [Alignment
        [{Agree, Affirm,\\
         Admit}]
      ]
      [Opposition
        [{Disagree, Refute,\\
         Deny}]
      ]
    ]
  ]
\end{forest}
\caption{Assertive Speech Act Taxonomy.}
\label{fig:speech-act-assertive}
\end{figure*}
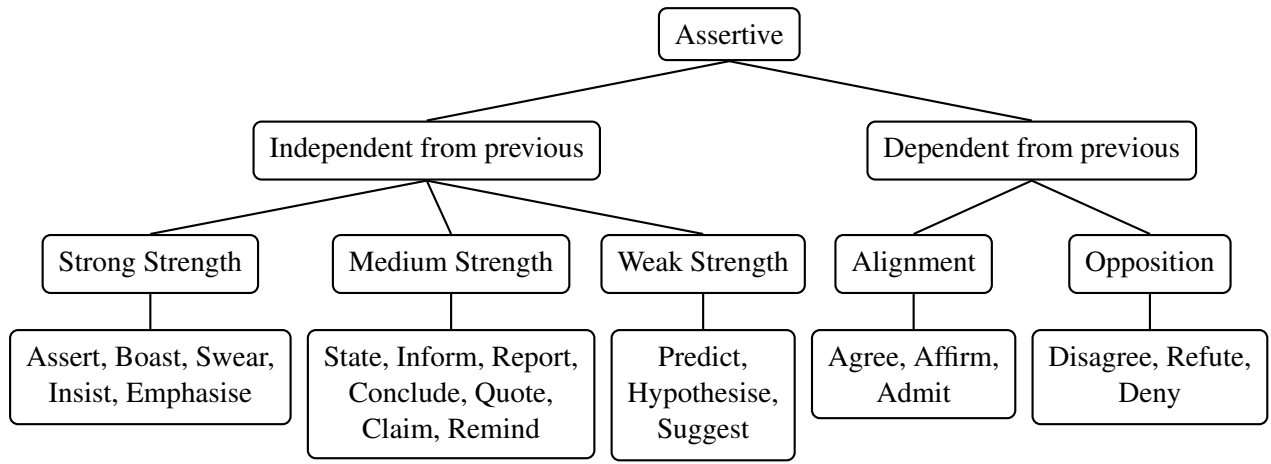

\begin{figure*}[htbp]
\centering
\begin{forest}
  for tree={
    node options={mynode},
    edge={thick, draw=black},
    parent anchor=south,
    child anchor=north,
    s sep=4mm, 
    l=1.5cm  
  }
  [Directive
    [Strong strength
      [{Order, Tell, Dare,\\ Question, Challenge}]
    ]
    [Medium strength
      [{Ask, Inquire}]
    ]
    [Weak strength
      [{Suggest, Recommend,\\ Beg, Pray}]
    ]
  ]
\end{forest}
\caption{Directive Speech Act Taxonomy}
\label{fig:speech-act-directive}
\end{figure*}
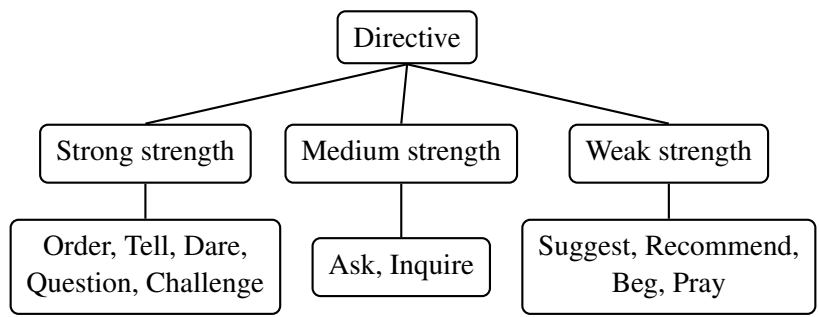

\begin{figure*}[htbp]
\centering
\begin{forest}
  for tree={
    node options={mynode},
    edge={thick, draw=black},
    parent anchor=south,
    child anchor=north,
    s sep=4mm,
    l=1.5cm  
  }
  [Commissive
    [Independent from previous
      [{Promise, Offer, Threaten, Warn}]
    ]
    [Dependent from previous
      [Accept]
      [Refuse]
    ]
  ]
\end{forest}
\caption{Commissive Speech Act Taxonomy}
\label{fig:speech-act-commissive}
\end{figure*}
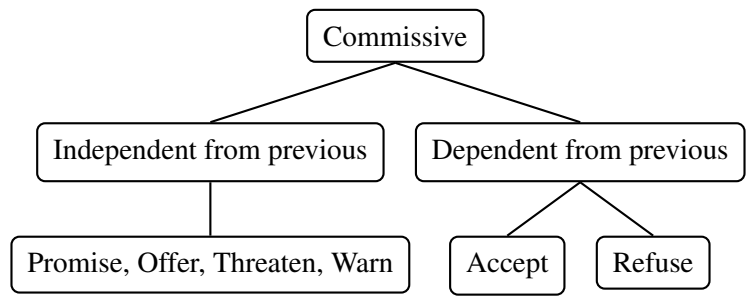

\begin{table*}[t]
\centering
\small
\begin{tabular}{lp{0.85\textwidth}}
\toprule
\textbf{Speech Act} & \textbf{Description} \\

\midrule
\textbf{Assertive} & commit the speaker in varying degrees to the truth of the expressed proposition\\
\hline
State & to say something clearly and carefully \\
Claim & to say that something is true or is a fact, although they cannot prove it and other people might not believe it \\
Report & to give a description of something or information about it \\
Inform & to say something they know and which they think the addressee should know and want to know \\
Suggest & to communicate or show an idea or feeling without stating it directly or giving proof \\
Hypothesise & to give a possible but not yet proved explanation for something \\
Predict & to say that an event or action will happen in the future, especially as a result of knowledge or experience \\
Insist & to say firmly, especially when others disagree with or oppose what they say \\
Swear & to promise or say firmly that they are telling the truth \\
Conclude & to judge or decide something after thinking carefully about it \\
Assert & to state an opinion forcefully \\
Affirm & to confirm something is true \\
Remind & to make someone aware of something forgotten or possibly forgotten, or to bring back a memory to someone \\
Admit & to agree that something is true, especially unwillingly \\
Agree & to say that they have the same opinion \\
Disagree & to say that they do not have the same opinion, idea, etc. \\
Refute & to say or prove that a person, statement, opinion, etc. is wrong or false \\
Emphasise & to state or show that something is especially important or deserves special attention; to make something more obvious \\
Deny & to say that something is not true, presents themselves as an authority on the subject or a person with special access to the relevant information \\
Boast & to speak too proudly or happily about what they have done or what they own \\
Quote & to repeat the words that someone else has said or written \\
 
\midrule
\textbf{Directive} & attempts of varying degrees by the speaker to get the hearer to do something\\
\hline
Order & express the assumption that the addressee has no choice but to do as they are told \\
Tell & express the assumption that the addressee will do as the speaker says \\
Suggest & express the message that it can be good if the addressee does something and thinks about doing it. There is no assumption that the addressee will necessarily do it. There is no component of 'I want you to do this'. \\
Recommend & speaker recognizing that the addressee wants to do something, informing the addressee of their appraisal of what it would be best to do, assuming the position of someone who is knowledgeable about the topic \\
Ask & speaker asking the addressee to do something, with uncertainty and a wish to know about whether addressee will do it \\
Beg & to ask someone to do something in an urgent way \\
Pray & to speak to a god either privately or in a religious ceremony in order to ask for something \\
Dare & to ask someone to do something that involves risk \\
Question & to express doubts about the value or truth of something, and wants people to think about it to know if it is right \\
Inquire & to ask for information \\
Challenge & to cause the addressee to do something or say that they are not able to do it. Speaker assumes that the addressee would say that they can do it, and assumes that it is difficult for the addressee to do \\

\midrule
\textbf{Commissive} & commit the speaker in varying degrees to some future course of action \\
\hline
Promise & to say that they will certainly do something \\
Offer & to declare themselves able to do something that can be good for the addressee and as being willing to do it. Speaker appears to be conscious of the possibilities that the addressee either may or may not want the speaker to do it (i.e. to either accept or decline the offer) \\
Threaten & to declare themselves able to do something that will be very bad for the addressee, to get the addressee to do something that the addressee doesn't want to do \\
Warn & speaker believes and expresses that if addressee do (or not do) something, something bad might befall the addressee. Want addressee to think about what they are going to do, to keep something in mind. \\
Refuse & to say that you will not do or accept something \\
Accept & to say yes to an offer or invitation \\
\bottomrule
\end{tabular}
\caption{Speech act taxonomy (part 1), including the speech act categories (Assertive, Directive, and Commissive) and classes, along with their brief description provided to LLM during extraction.}
\label{tab:speech-acts-1}
\end{table*}

\begin{table*}[t]
\centering
\small
\begin{tabular}{lp{0.85\textwidth}}
\toprule
\textbf{Speech Act} & \textbf{Description} \\

\midrule
\textbf{Expressive} & express the psychological state specified in the sincerity condition about a state of affairs specified in the propositional content \\
\hline
Apologize & to tell addressee that they are sorry for having done something that has caused problems or unhappiness for the addressee \\
Forgive & to express their wish to not think of addressee as of someone who did something bad to them, nor to feel something bad towards the addressee because of that \\
Thank & to express to addressee that they are pleased about or are grateful for something that the addressee have done \\
Complain & to say that something is wrong or not satisfactory. Speaker is focused on some bad effect they experienced as a result of someone else's actions, and they want someone to do something because of this. \\
Greet & to welcome someone with particular words or a particular action \\
Farewell & to say goodbye \\
Praise & to say something very good about someone else and to bring credit to the praised person \\
Criticize & to say something (merely) bad about someone else, to cause others to think something bad about the criticized person \\
Insult & to say or do something to someone that is rude or offensive \\
Congratulate & to praise someone and express that speaker approves of or is pleased about a special or unusual achievement \\
Deplore & to say or think that something is very bad \\
Wish & to express hope for another person's success or happiness or pleasure on a particular occasion; to express hope that speaker will get something or that something will happen; to express desire for some situation to be different from the one that exists \\

\midrule
\textbf{Other} & a generic class for speech acts not fitting into the classification \\
\hline
Other & a generic class for speech acts not fitting into the classification \\

\bottomrule
\end{tabular}
\caption{Speech act taxonomy part 2, including the speech act categories (Expressive and Other) and classes, along with their brief description provided to LLM during extraction.}
\label{tab:speech-acts-2}
\end{table*}

\begin{table*}[t]
\centering
\small
\begin{tabular}{p{0.2\textwidth} p{0.74\textwidth}}
\toprule
\textbf{Section} & \textbf{Text} \\
\midrule

\textbf{Role} &
You are an expert linguist and pragmatics specialist specializing in Speech Act Theory. You will be provided with a conversation context and a segmented target response. Your task is to classify each segment of the target response into specific illocutionary speech acts based on the provided taxonomy. \\

\midrule

\textbf{Speech Act Taxonomy \& Definitions} &
Illocutionary speech acts are the intended meaning of the utterance. The categories and speech acts within each category are defined below. \\
&
\texttt{\$\{SPEECH\_ACTS\}} \\

\midrule

\textbf{Instructions} &
1. Analyze: For each segment, analyze the conversation context, response context, and speech act definitions, to determine the illocutionary speech act that would be likely perceived by different types of listeners. Identify any signs of covert aggression, figurative language, indirectness, and violation of conversational maxims, that can make the literal and intended meaning of the text diverge.
    
\noindent2. Identify the Potential Intended Acts: For each segment, based on the analysis and potential factors diverging literal and intended meaning, assign the likely perceived intended speech acts and their corresponding categories using only the provided taxonomy.
\\

\midrule

\textbf{Expected Output Format} &
Return your analysis in the following JSON structure:

\begin{verbatim}
[
    {
        "segment index": 1,
        "text segment": "...",
        "analysis": "...",
        "potential factors diverging literal and intended meaning": "...",
        "intended speech acts": [category1-act1, ...]
    }
]
\end{verbatim}
\\

\midrule

\textbf{Conversation Context} &
The following is the history leading up to the target response. Use this to understand the intent, tone, and pragmatics.
\vspace{0.2cm}

\texttt{\$\{CONVERSATION\_HISTORY\}} \\

\midrule

\textbf{Target Response to Analyze} &
The target response is divided into segments, with each segment indexed and wrapped in \texttt{<segment n> ... </segment n>}. \\
&
Speaker: \texttt{\$\{SPEAKER\_NAME\}} \\
&
Response: \texttt{\$\{TARGET\_RESPONSE\}} \\

\bottomrule
\end{tabular}
\caption{LLM prompt structure for speech act extraction. The previous response to the target response, if available, is used as conversation history. Speaker names are anonymized into numerical identifiers (e.g., Speaker 1, Speaker 2).}
\label{tab:sa-extract-prompt}
\end{table*}

\end{document}